\documentclass[journal]{IEEEtran}
\usepackage{amsmath,amsfonts}
\usepackage{algorithmic}
\usepackage{algorithm}
\usepackage{array}
\usepackage[caption=false,font=normalsize,labelfont=sf,textfont=sf]{subfig}
\usepackage{textcomp}
\usepackage{stfloats}
\usepackage{url}
\usepackage{verbatim}
\usepackage{graphicx}
\usepackage{cite}
\usepackage{hyperref}

\usepackage{multirow}
\usepackage{booktabs}
\usepackage{tabularx}
\usepackage[table]{xcolor}
\begin{document}

\title{FSATFusion: Frequency-Spatial Attention Transformer for Infrared and Visible Image Fusion}

\author{Tianpei~Zhang, Jufeng~Zhao, Yiming~Zhu, Guangmang~Cui, Yuhan~Lyu}

\markboth{Journal of \LaTeX\ Class Files,~Vol.~14, No.~8, August~2021}%
{ \MakeLowercase{\textit{et al.}}: FSATFusion: Frequency-Spatial Attention Transformer for Infrared and Visible Image Fusion}


\maketitle
\begin{abstract}

The infrared and visible images fusion (IVIF) is receiving increasing attention from both the research community and industry due to its excellent results in downstream applications. Existing deep learning approaches often utilize convolutional neural networks to extract image features. However, the inherently capacity of convolution operations to capture global context can lead to information loss, thereby restricting fusion performance. To address this limitation, we propose an end-to-end fusion network named the Frequency-Spatial Attention Transformer Fusion Network (FSATFusion). The FSATFusion contains a frequency-spatial attention Transformer (FSAT) module designed to effectively capture discriminate features from source images. This FSAT module includes a frequency-spatial attention mechanism (FSAM) capable of extracting significant features from feature maps. Additionally, we propose an improved Transformer module (ITM) to enhance the ability to extract global context information of vanilla Transformer. We conducted both qualitative and quantitative comparative experiments, demonstrating the superior fusion quality and efficiency of FSATFusion compared to other state-of-the-art methods. Furthermore, our network was tested on two additional tasks without any modifications, to verify the excellent generalization capability of FSATFusion. Finally, the object detection experiment demonstrated the superiority of FSATFusion in downstream visual tasks. Our code is available at https://github.com/Lmmh058/FSATFusion.

\end{abstract}

\begin{IEEEkeywords}
Infrared and visible image fusion, Transformer, Deep learning, Attention mechanism.
\end{IEEEkeywords}

\section{Introduction}

\begin{figure}[!t]
\centering
\includegraphics[width=\columnwidth]{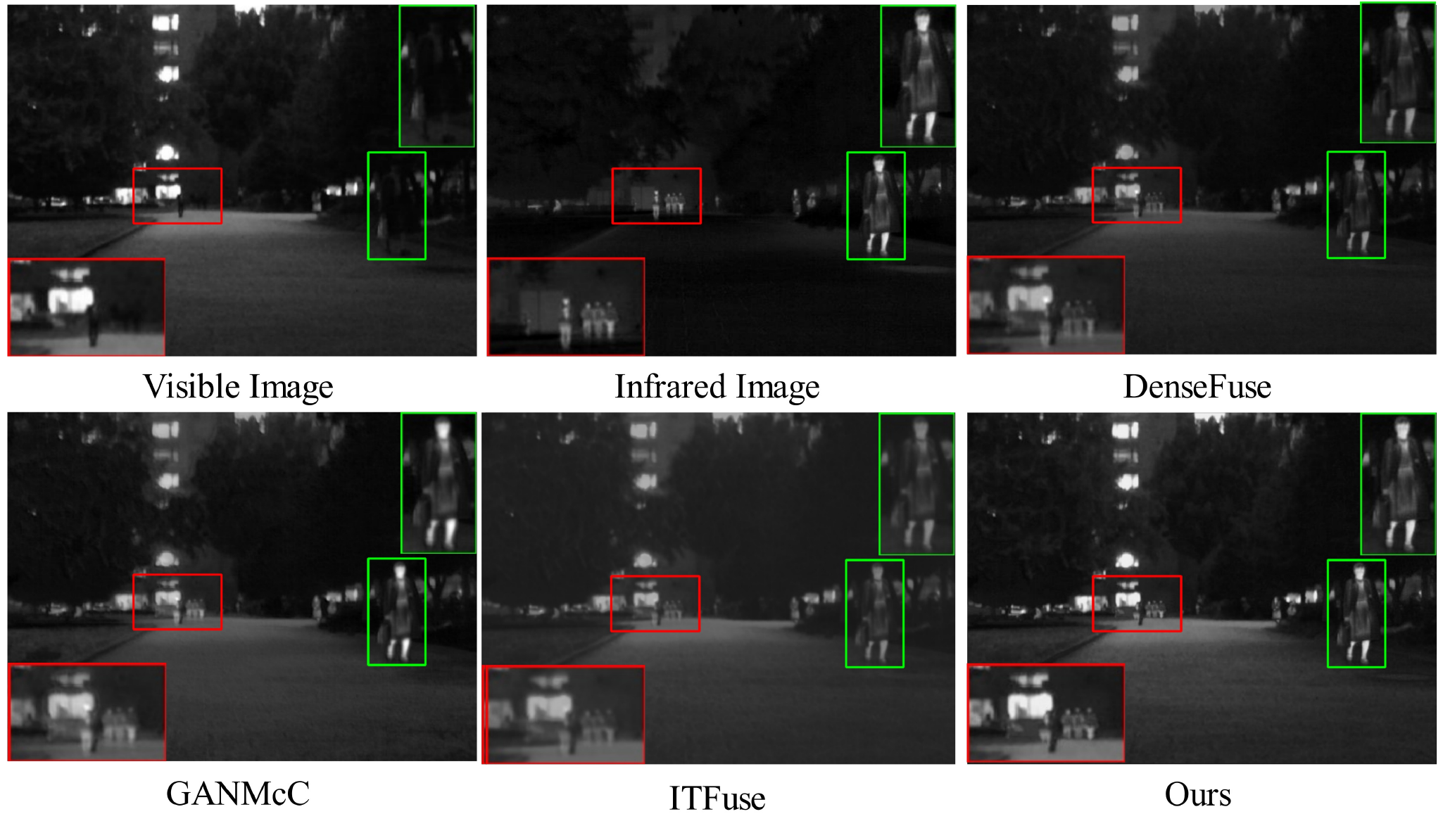}
\caption{Comparison of fusion results for a pair of images from MSRS dataset. These images are source images, the results of DenseFuse \cite{li2018densefuse}, GANMcC \cite{ma2020ganmcc}, ITFuse \cite{tang2024itfuse}, and our FSATFusion. Our result has more optimal detail description and prominent target percetion for a satisfactory visual effect. }
\label{Qualitative_first}
\end{figure}

% 1. 研究背景的意义
Image fusion is a pivotal technique in computer vision, with applications spanning image segmentation, object detection \cite{hu2025datransnet,xiao2024background}, recognition \cite{zhang2020object}, and tracking \cite{xiao2020review}. The inherent limitations of imaging technologies mean that a single image modality often fails to provide comprehensive information. Visible light images provide rich color and detail but are highly sensitive to brightness variations and occlusions. Conversely, infrared images capture thermal radiation even in harsh environments yet lack texture and color details. Consequently, infrared and visible image fusion (IVIF) combines the thermal information from infrared images with the texture details of visible images, yielding a composite image that is rich in information and aligned with human visual perception. This fused image enhances the performance of downstream computer vision tasks. As shown in Fig. \ref{Qualitative_first}, an example of infrared and visible image fusion demonstrates a composite image with complementary information and enhanced visual perception, which can facilitate streamlined tasks such as target detection in low-light conditions and improve the effectiveness of intelligent security systems.

% 图像融合是计算机视觉中的一项关键任务，具有广泛的应用，包括目标检测、识别和跟踪。由于成像技术的差异和局限性，单模态图像很难揭示全面的信息。可见光图像提供丰富的颜色和细节，但容易受到亮度变化和遮挡的影响。相反，红外图像即使在恶劣的环境中也能捕获热辐射，但缺乏纹理细节和颜色信息。因此，红外和可见光图像融合将红外图像的热细节与可见光图像的纹理细节相结合，创建了一个信息丰富、与人类视觉感知一致的合成图像。这种融合图像增强了下游计算机视觉任务的性能。红外和可见光图像融合的图示如图\ref{Qualitative_first}所示。具有互补信息和更好视觉感知的单个融合图像可以促进下游任务，如低光下的目标检测、智能安全等。

% 2. 所面临的挑战/关键技术瓶颈/关键科学问题

However, IVIF faces several challenges. Firstly, significant differences exist in the imaging principles and information content between infrared and visible light images, making it difficult to effectively combine important information from both sources. Secondly, both source images often contain useful features at multi-scale and multi-level dimensions, which requires comprehensive integration. Lastly, achieving a balance between high computational efficiency and satisfactory fusion performance in algorithm design represents another critical challenge. Consequently, current IVIF methods are plagued by limitations including low fusion efficiency, dependence on manually crafted complex fusion rules, and inadequate feature extraction from source images. 

% 3. 国内外研究现状
Over the past few decades, a multitude of image fusion methods have been proposed, broadly classified into two groups: traditional fusion methods \cite{wang2014fusion, selvaraj2020infrared} and deep-learning fusion models \cite{li2018densefuse, stfnet,wang2024uud, luo2024hbanet, huang2024fusiondiff,zhang2025exploring,zhang2025daaf}. The former utilize traditional image processing or statistical techniques to establish fusion frameworks. Meanwhile, the advent of deep learning has spurred the development of frameworks leveraging convolutional neural networks (CNNs) \cite{li2018densefuse, liu2023sgfusion, xu2020u2fusion} autoencoders (AE) \cite{li2021rfn,xu2022cufd}, and generative adversarial networks (GAN) \cite{ma2019fusiongan}, which have demonstrated remarkable effectiveness in various image fusion tasks \cite{apnet2024, lgabl2024}. In recent years, Vision Transformer \cite{dosovitskiy2020image} has achieved remarkable success in computer vision, has also been recently incorporated into image fusion.

% 4. 之前方法尚未解决的问题
% 然而，对于传统的融合框架，它们提取特征的能力有限，而且复杂的手动设计的融合策略阻碍了融合的性能和效率，进而影响融合结果。此外，现有的深度融合模型大多利用卷积神经网络（ConvNets）来进行特征提取，虽然表现出了良好的局部特征保存能力；然而，通过限制感受野和静态权重，它们构建远程关系和适应输入的能力受到限制。而最近基于Transformer的模型 \cite{tang2022ydtr, tang2023datfuse} 都没有考虑空间中不同模态的信息交互，导致互补信息挖掘不足。因此，尽可能充分利用不同模态图像的特征，并且构建远程关系是解决这一问题的关键因素。
A review of current image fusion methodologies reveals that traditional fusion frameworks suffer from limited feature extraction capabilities. Additionally, manually designed fusion strategies hinder both performance and efficiency, directly degrading fusion quality. Most existing deep fusion models rely on CNNs for feature extraction, which excel at preserving local features but are constrained in capturing long-range dependencies due to their limited receptive fields and static weight matrices. Recently, Transformer-based models \cite{tang2022ydtr, tang2023datfuse} have neglected the information interaction between spatial regions and frequency domains, resulting in insufficient complementary information extraction. Consequently, maximizing the exploitation of cross-modality features and modeling robust long-range dependencies are critical to addressing these challenges.

% 5. 本文创新 1
% 6. 本文创新 2
% 7. 本文创新 3 motivation

To address the persistent challenges of information loss in visual attention methods, inadequate global information extraction in CNN-based methods, and the suboptimal computational efficiency in Transformer-based methods, we proposed FSATFusion. Central to FSATFusion is the \textbf{frequency-spatial attention Transformer module (FSAT)}, which comprises two key components: the \textbf{Improved Transformer Module (ITM)} and the \textbf{Frequency-Spatial Attention Mechanism (FSAM)}. In the ITM, we integrate a novel lightweight attention mechanism termed Context Broadcast (CB) \cite{hyeon2023scratching} to enhance the Transformer's capability for global feature dependence. Simultaneously, FSAM identifies critical regions within the feature map across both the frequency and spatial domains. Frequency-domain analysis effectively captures global frequency features, while the spatial domain preserves significant positional information. Through applications of FSAM, we efficiently maintain the original information of the feature map while identifying its critical regions, thereby preventing information loss. In addition to FSAM, residual structures further enhance information retention. In conclusion, FSATFusion effectively integrates frequency-spatial attention with Transformer framework, achieving a harmonious balance between robust global information extraction and precise crucial local detail extraction. Its streamlined network architecture substantially enhances fusion efficiency.

% 补充说明
As illustrated in Fig. \ref{Qualitative_first}, we have compared FSATFusion with three state-of-the-art deep learning methods: DenseFuse \cite{li2018densefuse} (an AE-based method), GANMcC \cite{ma2020ganmcc} (a GAN-based method), and ITFuse \cite{tang2024itfuse} (a Transformer-based method). DenseFuse tends to generate overly dark outputs, resulting in the loss of visible information and an inability to accentuate infrared targets. GANMcC produces results that suffer from blurriness, characterized by washed-out textures and indistinct edges around salient targets. While ITFuse effectively retains visible details and offers a pleasing visual appearance, it falls short in terms of adequately highlighting infrared targets and introduces some noise. In contrast, FSATFusion not only maintains fine-grained visible details but also significantly enhances the saliency of infrared targets, producing a fused image with optimal brightness, contrast, and cross-modality consistency. This results in a more intuitive visual representation that aligns better with human vision perception and downstream task requirements.

% 8. 本文贡献
The key contributions of this paper are encapsulated in the following points:

\begin{enumerate}
\item{
We propose an end-to-end fusion network, FSATFusion, which incorporates a frequency-spatial attention Transformer, thereby eliminating the reliance on manually crafted fusion strategies. 
}
\item{
We present a frequency-spatial attention Transformer module that directs the network to concentrate on salient regions, significantly enhancing the global feature extraction performance of the standard Transformer.
}
\item{
Comprehensive experiments conducted on public \allowbreak databases demonstrate the
superiority of FSATFusion in terms of fusion performance and generalization ability.}
\item{
Comparative experiments in object detection illustrate that FSATFusion can be effectively integrated into downstream visual tasks.}
\end{enumerate}

The structure the rest paper is as follows: Sec. \ref{sec:relatedwork} provides an overview of related work. Sec. \ref{sec:method} presents the underlying principles and design of the network. Sec. \ref{sec:experiment} discusses the experimental results. The paper concludes with Sec. \ref{sec:conclusion}.

\section{Related work}
\label{sec:relatedwork}
\subsection{Infrared and Visible Image Fusion}
IVIF methods can be categorized into traditional and deep learning-based methods. Traditional methods encompass four primary categories: multi-scale transform, sparse representation, subspace-based, and saliency-based methods. Multi-scale transform methods \cite{burt1987laplacian,zhan2017infrared,li2016infrared,selvaraj2020infrared} decompose images into multi-scale representation, fuse these representation using predifined rules, and reconstruct the fused image through inverse transformation. Sparse representation methods \cite{wang2014fusion,lu2014infrared,yang2020infrared,zhang2023joint} extract image features by learning sparse codes from an over-complete dictionary, which includes dictionary construction, sparse coding, and the application of a fusion strategy. Subspace-based methods \cite{li2023infrared,lu2014novel,zhang2014multi} analyze the internal structure of images by mapping them from high-dimensional into low-dimensional spaces. Saliency-based \cite{cui2015detail,ma2017infrared,liu2022infrared} methods prioritize the preservation of salient features while minimizing noise, thus enhancing the visual quality of the fused output.
The selection of transformation techniques and fusion rules is pivotal, as they significantly influence the quality of fused images. Traditional methods often suffer from limitations in adaptive feature extraction and rely heavily on manually crafted fusion rules. Moreover, these approaches frequently employ uniform decomposition strategies across diverse modalities, disregarding the intrinsic characteristics of infrared and visible images. This oversight leads to inaccurate feature fusion and insufficient detail preservation in the fused results.

To address these limitations, numerous deep learning methods have been proposed. These methods can be categorized into three main categories: autoencoders (AE), convolutional neural networks (CNN), and generative adversarial networks (GAN). 

\begin{enumerate}
    \item \textbf{AE-based} methods comprise an encoder-decoder architecture designed to extract features from source images and reconstruct the fused image. Li \textit{et al.} \cite{li2021rfn} proposed RFN-Nest, which introduces a residual fusion network instead of manually designed fusion strategies. AE methods enhance performance by refining the network structure and incorporating attention mechanisms.

    \item \textbf{CNN-based} methods leverage their robust feature extraction capabilities and are extensively applied in IVIF; Liu \textit{et al.} \cite{liu2023sgfusion} proposed SGFusion, which utilizes saliency masks to emphasize important regions, thereby guiding the network's focus and maximizing the preservation of critical information; Xu \textit{et al.} \cite{xu2020u2fusion} developed U2Fusion framework, where different fusion tasks are designed to complement each other, enhancing overall fusion quality.

    \item \textbf{GAN-based} \cite{goodfellow2014generative} methods has gained popularity in IVIF due to their capacity to produce images rich in detail. Ma \textit{et al.} \cite{ma2019fusiongan} pioneered the integration of GANs into image fusion, with the generator crafting fused images that incorporate both visible light gradients and infrared intensity data. Subsequently, Ma \textit{et al.} \cite{ma2020ddcgan} introduced a dual-discriminator conditional GAN aimed at preserving features from both types of source images simultaneously.
\end{enumerate}

Despite their merits, these three categories of deep learning methods encounter distinct challenges: AE-based methods necessitate intricate, manually designed fusion strategies, which can be challenging to optimize during training; CNN-based methods are limited in their ability to extract global features due to the constraints of convolutional operations encounter difficulties in modeling long-range dependencies, resulting in potential information loss; and GAN-based methods are often subject to instability during the training phase.

\subsection{Transformer Architectural in Deep Learning}
The Transformer architecture, initially introduced in Natural Language Processing (NLP) \cite{vaswani2017attention}, has been widely utilized in various computer vision tasks, including image classification \cite{zhang2021transformer}, small target detection \cite{zhu2024towards}, and semantic segmentation \cite{lin2022ds}. Recently, numerous Transformer-based methods have been developed for IVIF. For instance, Tang \textit{et al.} \cite{tang2022ydtr} proposed a Y-shaped dynamic Transformer-based IVIF method that separately processes thermal radiation from infrared images and texture details from visible images through distinct branches, fusing them by a dynamic Transformer module before decoding the fused image. Additionally, Ma \textit{et al.} \cite{ma2022swinfusion} designed a generalized image fusion framework leveraging cross-domain long-range dependence,  integrating complementary information and facilitating global interactions through an attention-guided cross-domain module, and unifying multi-scene image fusion with considerations for structure preservation, detail preservation, and intensity control.

The aforementioned Transformer-based methods primarily focus on the spatial domain features of infrared images and texture details of visible images, as well as cross-domain long-range dependence. However, frequency features are often overlooked, leading to information loss and impeding the effective capture of cross-domain interactions between infrared and visible images. Furthermore, the computational inefficiency of Transformer models remains a significant challenge for optimization in IVIF. Our work distinguishes from previous studies in several key respects:

\begin{enumerate}
    \item \textbf{Comprehensive consideration of frequency and spatial domain information:} We propose a frequency-spatial attention Transformer module designed to significantly enhance global feature extraction capabilities.

    \item \textbf{Improved the conventional Transformer:} In the ITM, we integrate a lightweight attention mechanism known as Context Broadcast (CB), which enhances performance without substantially increasing computational complexity or model size.
\end{enumerate}

\subsection{Attention Mechanism in Vision Tasks}
The attention mechanism has emerged as an indispensable tool in computer vision, enabling models to emphasize key information by selectively emphasizing different aspects of features \cite{dai2024background}. In IVIF, the attention mechanism plays a vital role in enhancing salient features, capturing comprehensive global context, and thereby enhancing the overall quality of the fusion process \cite{hu2018squeeze, zhu2021dau, huang2023rdca, liu2022ssau}. Channel attention mechanisms enhance or suppresses features across different channels by assigning weights, thereby focusing on critical information \cite{hu2018squeeze}. The spatial attention mechanism focuses on the spatial dimension and enhances the feature representation of key regions. Wang \textit{et al.} \cite{wang2023attention} proposed the temporal-spatial attention module is constructed to enhance the both temporal and spatial information of images. Moreover, self-attention mechanism has demonstrated exceptional results in Transformer-based image fusion methods. Tang \textit{et al.} \cite{tang2023datfuse} propose DATFuse  utilizes a dual attention residual module for important feature extraction to improve the fusion performance of IVIF task. 

However, existing methods, whether it is channel \cite{hu2018squeeze} or spatial attention\cite{liu2022ssau}, have not fully considered frequency domain information in the fusion problem. For instance, texture information in infrared images may be obscured in frequency domain features. Relying solely on spatial domain attention mechanisms could result in information loss. Therefore, to achieve optimal fusion outcomes, it is imperative to consider information from both spatial and frequency domains comprehensively. Our work differs from our previous studies in these key aspects:

\begin{enumerate}
    \item \textbf{Comprehensive Analysis of Frequency Features:} Our module thoroughly analyzes the features in the frequency domain, encompassing high and low frequency information integration and their cross-modality interactions.

    \item \textbf{End-to-End Frequency-Spatial Attention:} We propose end-to-end modules that fully integrate attention mechanisms in both frequency and spatial domains.
\end{enumerate}

\section{Method}
\label{sec:method}
In this section, we will systematically present the components of our method. Initially, we introduce the overall network architecture in Sec. \ref{sec:3a}, followed by a detailed exposition of the proposed Frequency-Spatial Attention Transformer (FSAT) module in Sec. \ref{sec:3b}. Sec. \ref{sec:3c} will elucidate the formulation of the loss function that governs the training process.

\begin{figure*}[!t]
\centering
\includegraphics[width= \textwidth]{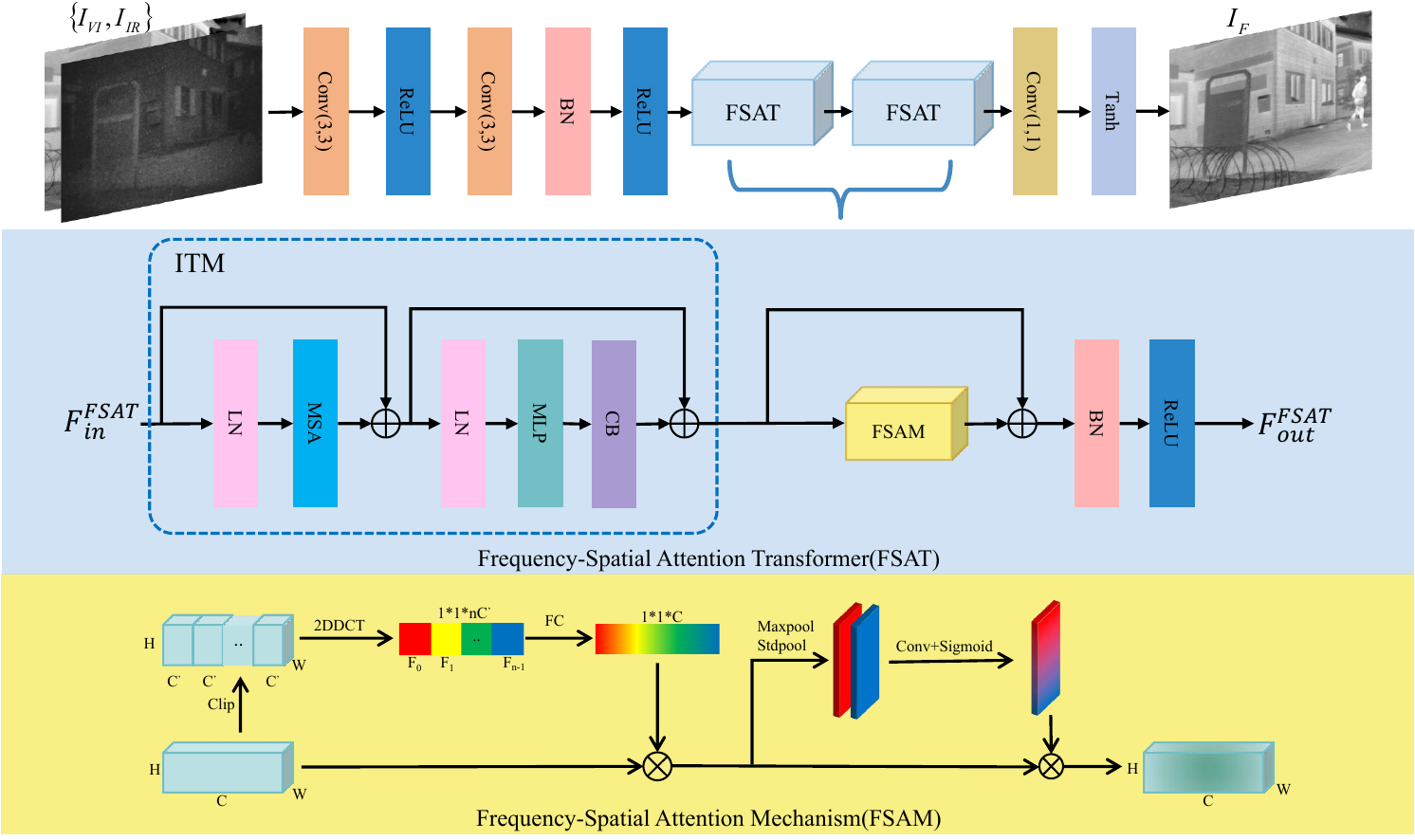}
\caption{A schematic overview of the proposed FSATFusion network is presented, highlighting the central components of the FSAT module. The architecture features two distinctively designed modules: the Improved Transformer Module (ITM), depicted in the blue box of Fig. 2, and the Frequency-Spatial Attention Mechanism (FSAM), shown in the yellow box of Fig. 2.}
\label{fig:Structure}
\end{figure*}

\subsection{Framework Overview}
\label{sec:3a}
The network architecture of FSATFusion is depicted in Fig. \ref{fig:Structure}, emphasizing the design of the Frequency-Spatial Attention Transformer (FSAT) module. Initially, the infrared image $I_{IR}$ and the visible image $I_{VI}$ are concatenated element-wise along the channel dimension to create a two-channel source image \(F_{in}\), mathematically expressed as:

\begin{equation}
\label{deqn_ex1}
F_{in}=concat(I_{IR},I_{VI})
\end{equation}

\noindent where \(concat(\cdot)\) represents the element-wise concatenation operation. This image undergoes shallow feature extraction through two convolutional layers, transforming the raw data into meaningful feature representations. This process can be expressed as follows:

\begin{equation}
\label{deqn_ex2}
F_{sf} = Re(BN(Conv_{3}^{C,C}(Re(Conv_{3}^{C_{in},C}(F_{in}))))))
\end{equation}

\noindent where \(F_{sf}\in R^{C\times H\times W}\) represents the shallow features obtained after convolution, \(Conv_{3}^{C_{in},C}(\cdot)\) represents the convolution operation with a \(3 \times 3\) kernel. Here, \(C_{in}\) and \(C\) correspond to the input and output channels, respectively. Within our proposed network, \(C_{in}\) correspond to the two concatenated source images, hence it is set to 2, while \(C\) is assigned to 16 to enable a more detailed feature representation. The function \(Re(\cdot)\) and \(BN(\cdot)\) represent the ReLU Activation Function Layer and Batch Normalization, respectively. The FSAT module is then tasked with extracting comprehensive global information and enhancing the salient features from these shallow features, as articulated by the equation:

\begin{equation}
\label{deqn_ex3}
F_{FSAT}^{n}=FSAT^{C}(F_{n})
\end{equation}

\noindent where \(F_{n}\) are defined as:

\begin{equation}
\label{deqn_ex4}
{F_{n}} = 
\begin{cases}
F_{sf}, &{n=1},\\
{F_{FSAT}^{n-1},}&{n>1.}
\end{cases}
\end{equation}

\noindent where \(F_{FSAT}^{n} \in R^{C \times H \times W}\) denotes the feature map obtained by the \(nth\) FSAT module, and \(FSAT^{C}(\cdot)\) denotes the FSAT module with input and output channels both equal to \(C\). The specific operational principles of this module are detailed in Sec. \ref{sec:3b}. Finally, a convolutional layer with a kernel size of \(1 \times 1\) is employed for feature integration and image reconstruction, as delineated by:

\begin{equation}
\label{deqn_ex5}
I_{F}=Tanh(Conv_{1}^{C,1}(F_{FSAT}^{n}))
\end{equation}

\noindent where \(I_{F}\) represents the fused image, \(Tanh(\cdot)\) represents the Tanh activation function, and \(Conv_{1}^{C,1}\) represents the convolution operation with a \(1 \times 1\) kernel, where the input and output channels are \(C\) and 1, respectively.

This architecture synergistically combines CNNs for local feature extraction and Transformers for global context modeling, enabling FSATFusion to preserve fine-grained details, enhance cross-modality consistency, and highlight salient thermal targets in fused results.

\subsection{Frequency-Spatial Attention Transformer}
\label{sec:3b}
The design principles of the Frequency-Spatial attention Transformer (FSAT) module are integral to our network architecture. The FSAT module addresses the limitations of shallow feature extraction by integrating global context modeling (via ITM) and multi-domain attention (via FSAM). As shown in Fig. \ref{fig:Structure}, the feature map then proceeds to the FSAT module. Within this module, it initially passes through the Improved Transformer Module (ITM) to capture long-range dependencies and acquire global features. Afterwards the Frequency-Spatial Attention Module (FSAM) identifies critical regions within the feature map, considering both frequency and spatial perspectives, thereby effectively preserving rich global frequency features and significant spatial position information. 

\textbf{Improved Transformer Module:}Initially, the input features \(F_{in} \in R^{C \times H \times W}\) are processed by the ITM, the structure of which is designed based on \cite{liu2021swin}. The ITM first utilizes the PatchEmbedding operation to partition the input feature map into \(\frac{H}{M} \times \frac{W}{M} \) patches of size \(M^{2} \times C\), and subsequently calculates their self-attention. For the input matrix \(X \in R^{M^{2} \times C}\) to the self-attention mechanism, it undergoes linear transformation by three sets of matrices with trainable weights to generate the matrices \(Q\), \(K\), and \(V\), termed the query, key, and value matrices, respectively:

\begin{equation}
\label{deqn_ex6}
\begin{aligned}
Q &= XW^{Q} \\
K &= XW^{K} \\
V &= XW^{V}
\end{aligned}
\end{equation}

\noindent where \(W^{Q}\), \(W^{K}\), and \(W^{V}\) are the weight matrices of linear transformations. Attention metrics are calculated after obtaining the  \(Q\),  \(K\), and  \(V\) matrices:

\begin{equation}
\label{deqn_ex7}
Attention(Q,K,V) = softmax(\frac{QK^{T}}{\sqrt{d_{k}}})V
\end{equation}

\noindent where \(d_{k}\) is the scaling factor to prevent the dot product value from being too large and \(softmax(\cdot)\) denotes the normalized exponential function. The output is then connected to the input with a residual structure and non-linearly transformed through a Multi-Layer Perceptron (MLP) to enhance the feature representation's diversity and complexity.

According to \cite{hyeon2023scratching}, the Vision Transformer (ViT) tends to learn dense attention maps, but faces challenges with gradient descent optimization. To surmount this problem, we insert Context Broadcast (CB) subsequent to the MLP layer. This addition facilitates the dense interactions required by ViT for more efficient learning. For the CB layer with input \(X \in R^{N \times D}\), this process can be articulated  with the following equation:

\begin{equation}
\label{deqn_ex8}
x_{CB}^{n} = \frac{1}{2}x^{n}+\frac{1}{2N} \sum_{i=1}^{N}x^{i}
\end{equation}

\noindent where \(x^{n}\) represents the \(nth\) D-dimensional vector in \(X\).

\textbf{Frequency Spatial Attention Module:} Following the extraction of global features in the ITM, 
the feature map \(F_{ITM} \in R^{C \times H \times W}\) is then input into FSAM to determine the significance of features and their spatial locations. The CBAM \cite{woo2018cbam} argue that an exclusive reliance on global average pooling for the computation of attention maps results in the loss of information and the extraction of suboptimal features. Therefore, drawing inspiration from \cite{qin2021fcanet}, We employ two-dimensional discrete cosine transform (2D-DCT) to extract the diverse frequency components of the feature map. These components then serve as weight coefficients for the construction of the weight map, ensuring that information from each frequency component is preserved. The 2D-DCT can be mathematically expressed as:

\begin{equation}
\label{deqn_ex9}
f_{a,b} = \lambda(a)\lambda(b)\sum_{h=0}^{H-1}\sum_{w=0}^{W-1}C_{h,w}^{a,b}x_{h,w}
\end{equation}

\noindent where \(\lambda(a)\) (same as \(\lambda(b)\)) are defined as:
\begin{equation}
\label{deqn_ex10}
{\lambda(a)} =
\begin{cases}
\sqrt{\frac{1}{N}}, &{a=0},\\
{\sqrt{\frac{2}{N}},}&{otherwise.}
\end{cases}
\end{equation}

\noindent in ({\ref{deqn_ex9}}), \(\lambda(a)\) are the normalization coefficients of 2D-DCT, \(f\) denotes the 2D-DCT spectrum where \(a\in (0,H-1)\) and \(b \in (0,W-1)\), \(x\in R^{H \times W}\) are the inputs, the coefficients \(C_{h,w}^{a,b}\) are denoted as follows:

\begin{equation}
\label{deqn_ex11}
C_{h,w}^{a,b} = cos(\frac{\pi a}{H}(h+\frac{1}{2}))cos(\frac{\pi b}{W}(w+\frac{1}{2}))
\end{equation}

\noindent from ({\ref{deqn_ex9}}) and ({\ref{deqn_ex11}}), when \((a,b)=(0,0)\), we get \(C_{0,0}^{h,w}=1\) and thus:

\begin{equation}
\label{deqn_ex12}
f_{a,b} = \lambda(a)\lambda(b) \times HW \times avgpool(x)
\end{equation}

\noindent as seen in ({\ref{deqn_ex12}}), 
the global average pooling operation is directly proportional to the lowest frequency component of 2D-DCT, which also demonstrates that relying solely on global average pooling to compute attention weight maps leads to substantial information loss. This is because other frequency information within the feature map is discarded during the calculation, undoubtedly impeding the retention of information in the feature map. Therefore, we divide the feature map \(F_{ITM}\in R^{C \times H \times W}\) into \(N\) equal parts along the channel dimension \(F_{ITM,n}\in R^{C' \times H \times W}, n \in (0,1,\cdot\cdot\cdot,N-1)\), where \(C=N \times C'\), and then we assign different 2D-DCT frequency components to each different part instead of global average pooling to avoid information loss from other frequencies:

\begin{equation}
\label{deqn_ex13}
f^{n} = \lambda(a_n)\lambda(b_n) \sum_{h=0}^{H-1} \sum_{w=0}^{W-1} C_{h,w}^{a_{n},b_{n}}F_{ITM,n}^{h,w}
\end{equation}

\noindent after that we obtain the attention weight maps through fully connected layer and Sigmoid activation function respectively. The computation of the frequency attention weight process can be written as:

\begin{equation}
\label{deqn_ex14}
Att_{f}=Sigmoid(FC(concat(f^{0},f^{1},\cdot\cdot\cdot,f^{N-1})))
\end{equation}

\noindent where \(Att_{f}\) is the normalized attention weight map obtained after frequency attention, \(Sigmoid(\cdot)\) is the Sigmoid activation function, \(FC(\cdot)\) denotes the fully connected layer, and \(concat(\cdot)\) denotes the concatenation operation. The final step involves performing an element-wise multiplication of the feature map and the attention weight map \(F_{f} = Att_{f} * F_{ITM}\). This operation yields the output feature map for frequency attention.

Spatial attention is distinct from frequency attention in that it focus on the positional information of salient features. Accordingly, we have designed spatial attention to enhance the capacity to pinpoint focal information, thereby constituting FSAM along with frequency attention. Given that applying pooling operation along the channel axis have been shown to be effectively emphasize information regions \cite{zagoruyko2016paying}, we implement both maximum and standard deviation pooling along the channel direction on the feature maps derived from frequency attention \(F_{f}\in R^{C \times H \times W}\) to produce two distinct spatial information descriptors \(F_{pool}\in R^{d \times H \times W}\):

\begin{equation}
\label{deqn_ex15}
F_{pool} = concat(Maxpool(F_{f}),Stdpool(F_{f}))
\end{equation}

\noindent where \(Maxpool(\cdot)\) and \(Stdpool(\cdot)\) denote the max pooling and standard deviation pooling operations, respectively. Given that there are two distinct pooling processes, we set \(d=2\). To derive the spatial attention weight map, we use the convolution operation on the concatenated tensor, followed by the application of the Sigmoid activation function. The comprehensive process can be expressed by the following equation:

\begin{equation}
\label{deqn_ex16}
Att_{s} = Sigmoid(Conv_{7}^{2,1}(F_{pool}))
\end{equation}

\noindent where \(Att_{s}\) is the normalized attention weight map, \(Conv_{7}^{2,1}(\cdot)\) denotes the convolution operation with a convolution kernel of \(7 \times 7\) and the number of input and output channels are 2 and 1, respectively. The output feature map \(F_{s} = Att_{s} * F_{f}\), representing the output of the spatial attention, is subsequently computed.

\subsection{Loss Function}
\label{sec:3c}
In the training phase, our objective is to maximize the complementary information from the two source images, while preserving their structural and textural details. This is achieved through a composite loss function that includes pixel loss, texture loss, and SSIM loss, each addressing distinct aspects of fusion quality.

To capture as much information as possible from the source images, we employ mean square error between the fusion output and the source images, which term pixel loss. This approach quantifies pixel-level differences, ensuring that the fused image maintains appropriate intensity characteristics. The pixel loss is mathematically defined as:

\begin{equation}
\label{deqn_ex17}
\mathcal{L}_{pixel} = \frac{1}{HW}\sum^{H}_{h=1}\sum^{W}_{w-1}[(i_{h,w}^{F}-i_{h,w}^{VI})^{2}+(i_{h,w}^{F}-i_{h,w}^{IR})^{2}]
\end{equation}

\noindent where \(\mathcal{L}_{pixel}\) represents the pixel loss, \(H\) and \(W\) denote the height and width of the image, and \(i_{h,w}^{F}\), \(i_{h,w}^{VI}\), \(i_{h,w}^{IR}\) denote the pixel values at \((h,w)\) for the fused image, visible source image, and the infrared source image, respectively.

Texture loss is employed to enhance the incorporation of textural details from the source images into the fused result. Although infrared images contain fewer texture details than visible images, they still provide essential edge information. We calculate the texture loss by comparing the gradient magnitudes of the source images, selecting the greater values, and then assessing the disparity between these and the gradients of the fused image. The texture loss is formulated as:

\begin{equation}
\label{deqn_ex18}
\mathcal{L}_{texture} = \|\nabla I_{F} - Max(\nabla I_{VI}, \nabla I_{IR})\|_{2}
\end{equation}

\noindent where \(\mathcal{L}_{texture}\) is the texture loss, \(\| \cdot \|_{2}\) denotes \(l_{2}\)-norm, \(\nabla\) denotes Sobel gradient operator, \(Max(\cdot)\) is the operation of taking the maximum value, and  \(I_{F}\), \(I_{VI}\), and \(I_{IR}\) are the fused image, visible light image and infrared image, respectively.

Structural Similarity (SSIM) evaluates brightness, contrast, and structural consistency between images. In IVIF, low-light visible images can degrade fused output quality, reducing target saliency. To mitigate this, we compute the maximum pixel values from the two source images, the SSIM loss is formulated as:

\begin{equation}
\label{deqn_ex19}
I_{Max}(h,w)=Max(I_{VI}(h,w),I_{IR}(h,w))
\end{equation}

\noindent where \(w \in(0,W-1)\), \(h\in(0,H-1)\).

The SSIM loss function is then calculated between the fused output and this maximum value, aiming to minimize the information loss due to source image degradation and enhance the visual quality of the fused image, the SSIM loss function is defined as:

\begin{equation}
\label{deqn_ex20}
\mathcal{L}_{SSIM} = 1-SSIM(I_{F},I_{Max})
\end{equation}

\noindent where \(\mathcal{L}_{SSIM}\) denotes the structural similarity loss function and \(SSIM(\cdot)\) denotes the operation to calculate the SSIM value.

In conclusion, the overall loss function of FSATFusion is given by:

\begin{equation}
\label{deqn_ex21}
\mathcal{L} = \alpha \mathcal{L}_{pixel} + \beta \mathcal{L}_{texture} +\gamma \mathcal{L}_{SSIM}
\end{equation}

\noindent where \(\alpha\), \(\beta\) and \(\gamma\) are the weight coefficients of the loss function.

\begin{table*}[!t]
  \setlength{\abovecaptionskip}{0cm}  
  \renewcommand\arraystretch{1.2}
  \footnotesize
  \centering
  \vspace{-1\baselineskip}
\caption{CONFIGURATIONS FOR ALL COMPARATIVE EXPERIMENTS}
\vspace{-1\baselineskip}
\label{tab:parameters}
\tabcolsep=0.3cm
\renewcommand\arraystretch{1.2}
\begin{center}
 \resizebox{\textwidth}{!}{
\begin{tabular}{ll}
\toprule[1pt]
Methods(\textit{Source \& Year \& Type})         & Key parameters configurations   \\ \hline
\multicolumn{2}{l}{\textit{Traditional Infrared and Visible Image Fusion Methods}}  \\ \hline
CBF \cite{shreyamsha2015image} \textit{2015, Multi-scale-based} & Neighborhood Window Size: 11 $\times$ 11, $\sigma_{s} = 1.8, \sigma_{r} = 25$ \\
LatLRR \cite{li2018infrared} \textit{2018,  Saliency-based} & $\omega_{1} = \omega_{2} = 0.5, s_{1} = s_{2} = 1, \lambda = 0.8$ \\ 
\hline
\multicolumn{2}{l}{\textit{Deep-Learning-based Infrared and Visible Image Fusion Methods}}  \\ 
\hline
DenseFuse \cite{li2018densefuse} \textit{TIP'2018, AE-based} & Patch Size: 256 $\times$ 256, $\lambda = \{1,10,100,1000\}$ \\
GANMcC \cite{ma2020ganmcc} \textit{TIM'2020, GAN-based} & Patch Size: 120 $\times$ 120, $b = 32, p = \frac{1}{2}, M = 10, \gamma = 100, \beta_{1} = 1, \beta_{2} = 5, \beta_{3} = 4, \beta_{4} = 0.3$ \\ 
PMGI \cite{zhang2020rethinking} \textit{AAAI'2020, CNN-based} &    Patch size: \{60 $\times$ 60, 120 $\times$ 120\}\\  
CSF \cite{xu2021classification} \textit{TCI'2021, AE-based} & Patch Size: 128 $\times$ 128, $\lambda = 25, \sigma = 0.0001, L = 30, \alpha=0.7$ \\ 
RFN-Nest \cite{li2021rfn} \textit{INFFUS'2021, AE-based} & Patch Size: 256 $\times$ 256, $\lambda = 100, M=4$ \\           
SwinFusion \cite{ma2022swinfusion} \textit{JAS'2022, Transformer-based}    &  Patch size: 128 $\times$ 128, $\lambda_{1} = 10, \lambda_{2} = 20, \lambda_{3} = 20$ \\ 
DATFuse \cite{tang2023datfuse} \textit{TCSVT'2023, Transformer-based} & Patch Size: 120 $\times$ 120, $C = 16, R = 4, \alpha=1, \lambda = 100, \gamma = 10$ \\ 
MDA \cite{yang2024mda} \textit{NEUCOM'2024, CNN-based}    &  Patch size: 192 $\times$ 192, $\alpha = 1e-8, \beta = 1e5, \gamma = 2 ,\eta = 0.02, \sigma = 0.167, \xi = 20, c = 3 \times 10^3$ \\ 
ITFuse \cite{tang2024itfuse} \textit{PR'2024, Transformer-based}    &  Patch size: 120 $\times$ 120, $N = 3, C = 16, R = 4, L = 3, \epsilon = 0.001, \alpha = \beta = 1, \gamma = \lambda = 4$ \\ 

\rowcolor[rgb]{0.9,0.9,0.9}$\star$ \textbf{FSATFusion(Ours), Transformer-based} & Patch size: 224 $\times$ 224, $\alpha = 1, \beta = 10, \gamma = 100$ \\
\bottomrule[1pt]
\end{tabular}}
\end{center}
\end{table*}

\begin{figure}[!t]
\centering
\includegraphics[width=\columnwidth]{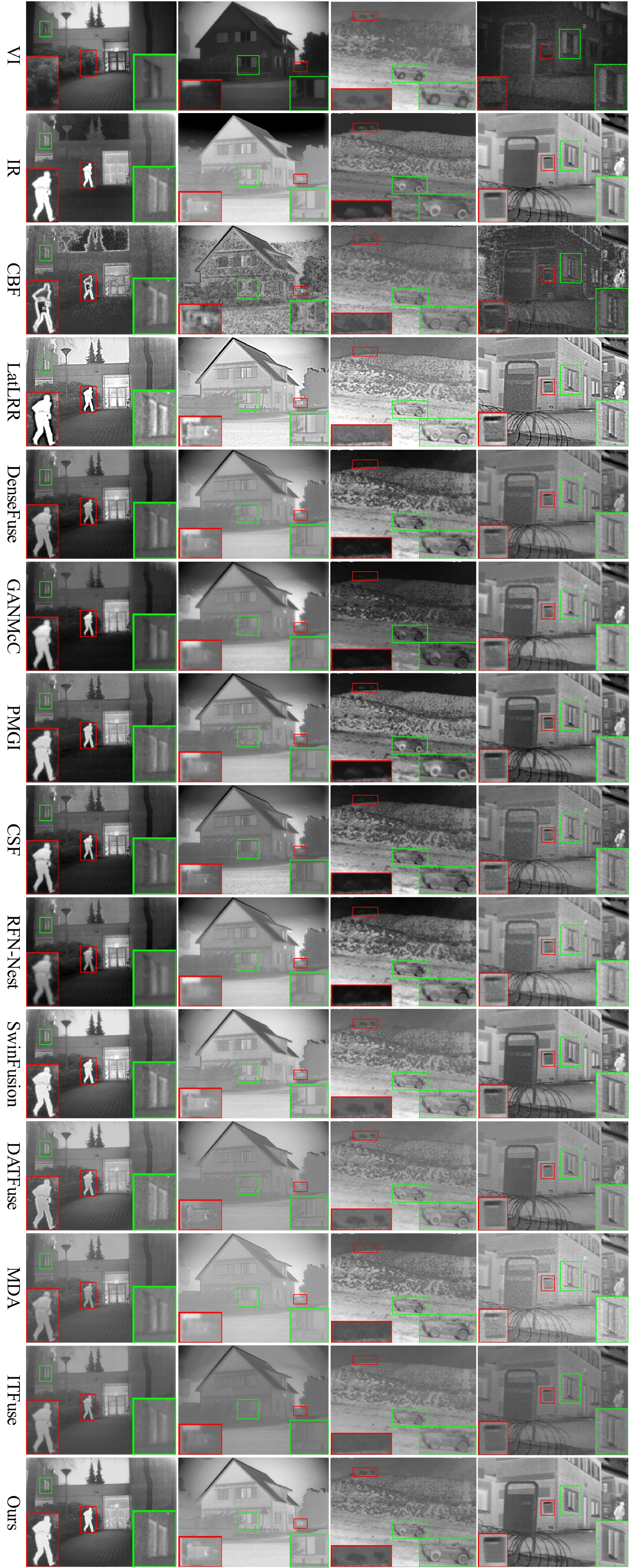}
\caption{Qualitative comparison of FSATFusion and eleven comparative methods on TNO dataset. For clearer comparison, we highlighted and enlarged two important parts in each figure, marked with red and green boxes respectively.}
\label{Qualitative_TNO}
\end{figure}

\begin{figure}[!t]
\centering
\includegraphics[width=0.97\columnwidth]{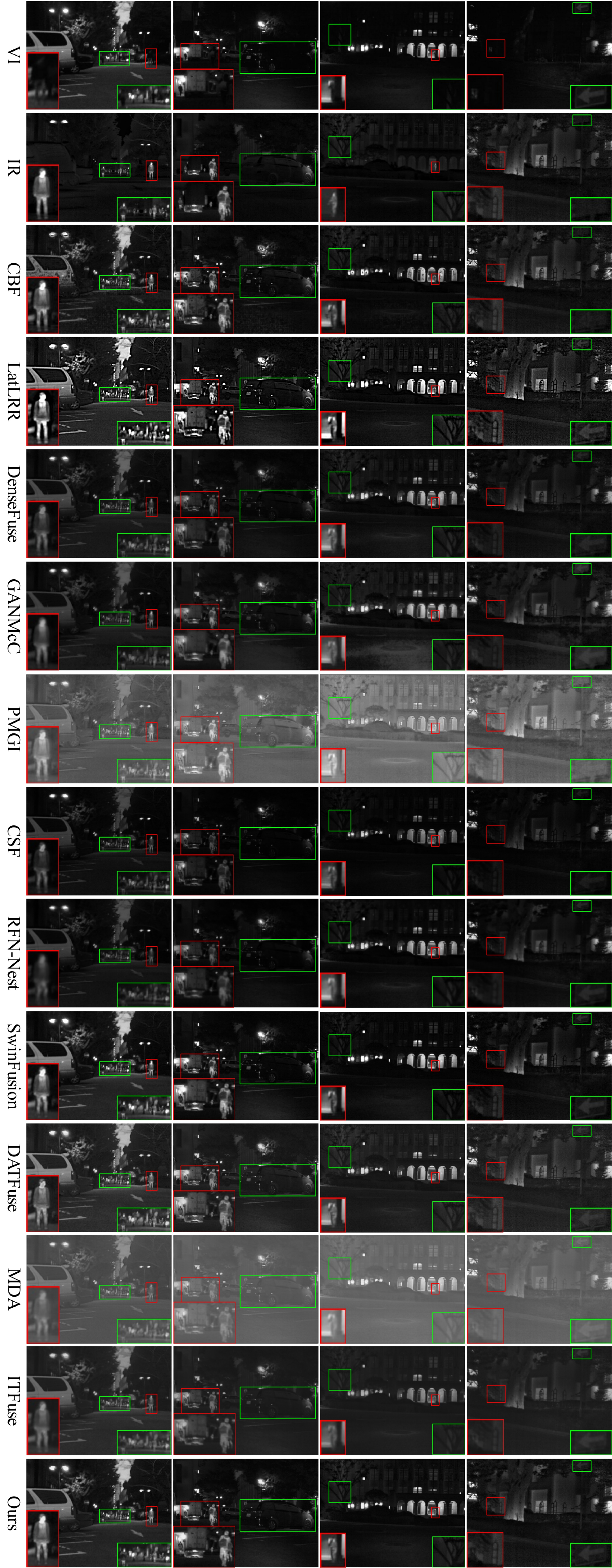}
\caption{Qualitative comparison of FSATFusion and eleven comparison methods on MSRS dataset. For clearer comparison, we highlighted and enlarged two important parts in each figure, marked with red and green boxes respectively.}
\label{Qualitative_MSRS}
\end{figure}

\section{Experiments and Analysis}
\label{sec:experiment}
This section provides an overview of the structure of our experimental and analysis. Sec. \ref{sec:4a} describes the datasets utilized in our experiments and outlines the training parameters. Sec. \ref{sec:4b}  discusses the comparative methods and the quantitative evaluation metrics that were employed. Sec. \ref{sec:4c} presents both quantitative and qualitative analyses of the proposed FSATFusion compared to other state-of-the-art methods on the infrared and grayscale visible image fusion task using the TNO and MSRS datasets. Sec. \ref{sec:4d} elaborates on the ablation studies conducted to assess the network structure and the loss function. Sec. \ref{sec:4e} exhibits FSATFusion's generalization capability through experiments involving the infrared and RGB image fusion, as well as near-infrared and RGB images fusion. Sec. \ref{sec:4f} evaluates the model's operational efficiency across various datasets, underscoring the efficiency of our method in image fusion tasks. Finally, Sec. \ref{sec:4g} presents comparative experiments employing fusion algorithms for downstream tasks, highlighting the compatibility of our method with subsequent visual tasks.

\subsection{Datasets and Experimental Setup}
\label{sec:4a}
The images utilized for training and testing were obtained from publicly accessible datasets. Our FSATFusion model was trained using 12,025 pairs of images from the LLVIP dataset \cite{jia2021llvip}. For evaluating network performance, we assembled test sets comprising 41 pairs from TNO \cite{toet2017tno}, 81 pairs from MSRS \cite{tang2022piafusion}, 75 pairs from RoadScene \cite{xu2020fusiondn}, and 88 pairs from the RGB-NIR \cite{brown2011multi} datasets.

During the training phase, the dimensions of the input visible and infrared images were resized to \(224 \times 224\), with a batch size of 32. The model training was facilitated by the Adam optimizer, over 50 epoch with an initial learning rate of 1e-3. A cosine annealing learning rate schedule was implemented to dynamically adjust the learning rate, thereby aiding for rapid convergence to an optimal solution. The weights \(\alpha\), \(\beta\), and \(\gamma\) of the loss function are set to 1, 10, and 100, respectively. The network model was implemented using the PyTorch framework and the experiments were executed on platforms equipped with an AMD EPYC 9754 and an NVIDIA GeForce RTX 4090D.

\subsection{Compared Method and Quantitative Evaluation Index}
\label{sec:4b}

To demonstrate the efficacy of FSATFusion, a comparative analysis was conducted with two exemplary traditional methods—CBF \cite{shreyamsha2015image} and LatLRR \cite{li2018infrared}—as well as nine leading deep learning approaches: DenseFuse \cite{li2018densefuse}, PMGI \cite{zhang2020rethinking}, CSF \cite{xu2021classification}, RFN-Nest \cite{li2021rfn}, GANMcC \cite{ma2020ganmcc}, SwinFusion \cite{ma2022swinfusion}, DATFuse \cite{tang2023datfuse}, MDA \cite{yang2024mda}, and ITFuse \cite{tang2024itfuse}. In conclusion, eleven comparative methods were assessed against FSATFusion, with SwinFusion, DATFuse, and ITFuse noted as the most sophisticated Transformer-based deep learning algorithms.

Quantitative comparative experiments were executed by assessing the fusion quality across four dimensions: information theory, image features, structural similarity, and human perception. We adopted two information theory-based metrics: Mutual information(\(MI\)) \cite{qu2002information}, Nonlinear correlation information entropy(\(NCIE\)) \cite{wang2005nonlinear}; two image feature-based metrics: gradient-based similarity measurement(\(Q_{abf}\)) \cite{xydeas2000objective}, phase congruency(\(Q_{p}\)) \cite{zhao2007performance}; one structural similarity-based metrics: Yang’s metric(\(Q_{y}\)) \cite{li2008novel} and one human perception inspired metric: visual information fidelity(\(VIF\)) \cite{han2013new}. A total of six evaluation indicators are used to objectively evaluate and compare the fusion results. \(MI\) measures how much information from the source images is retained in the fused result; \(NCIE\) assesses the nonlinear correlation between the fused image and the source images; \(Q_{abf}\) evaluates the quality of visual information in the fused image based on edge information; \(Q_{P}\) reflects the ability of fused images to preserve and integrate input features through phase consistency;  \(Q_{y}\) assesses the structural similarity of the fused image to the source images; \(VIF\) evaluates the visual information fidelity of the fused image to the source images. Superior fusion performance and outcomes are indicated by higher values of these metrics.

\begin{figure*}[!t]
\centering
\includegraphics[width=\textwidth]{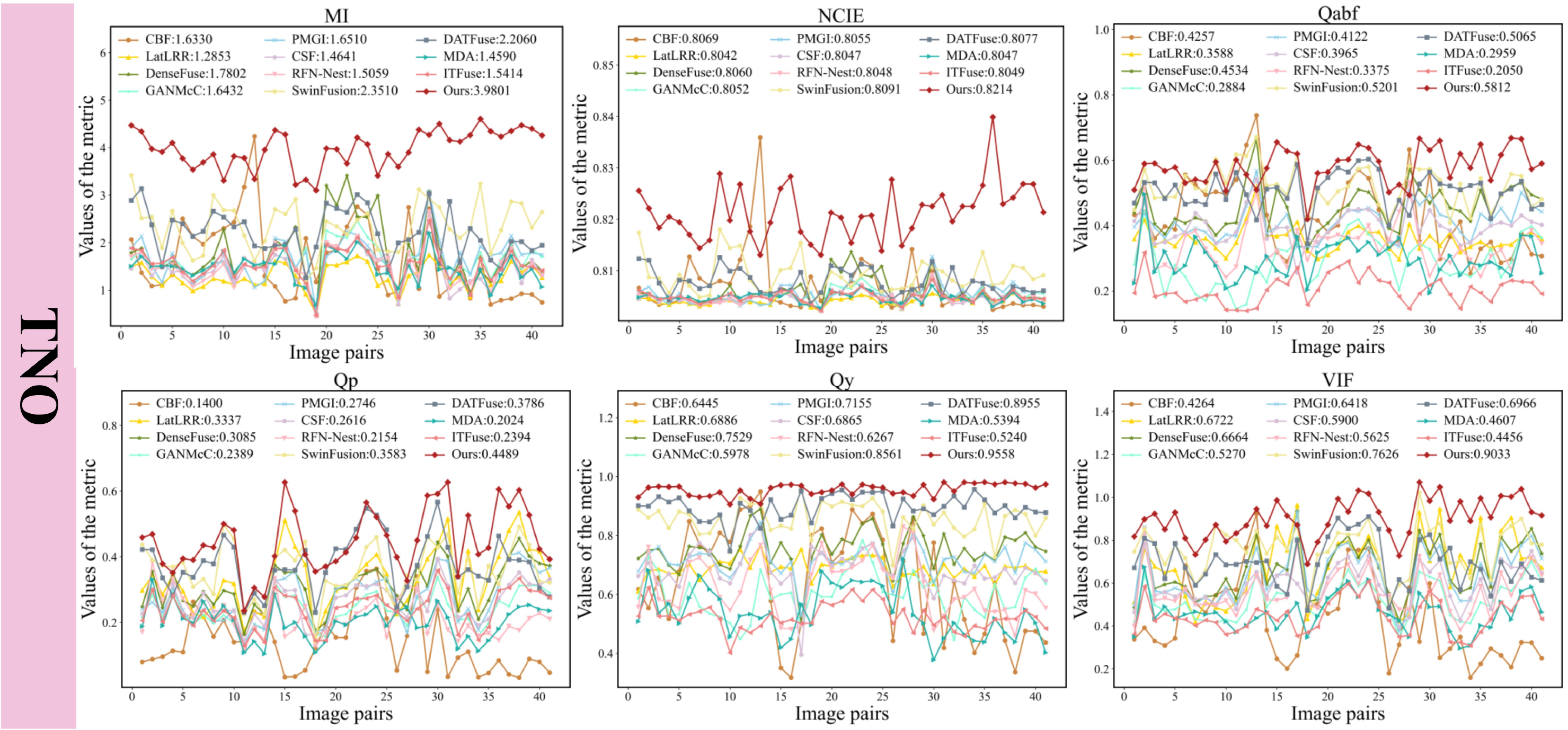}
\caption{Quantitative comparison of FSATFusion and eleven comparison methods on TNO dataset. Our method is represented by the red line. Mean values of each method are shown in each legend. For the six metrics used for comparison, a higher value indicates superior performance.}
\label{Quantitative_TNO}
\end{figure*}

\begin{figure*}[!t]
\centering
\includegraphics[width=\textwidth]{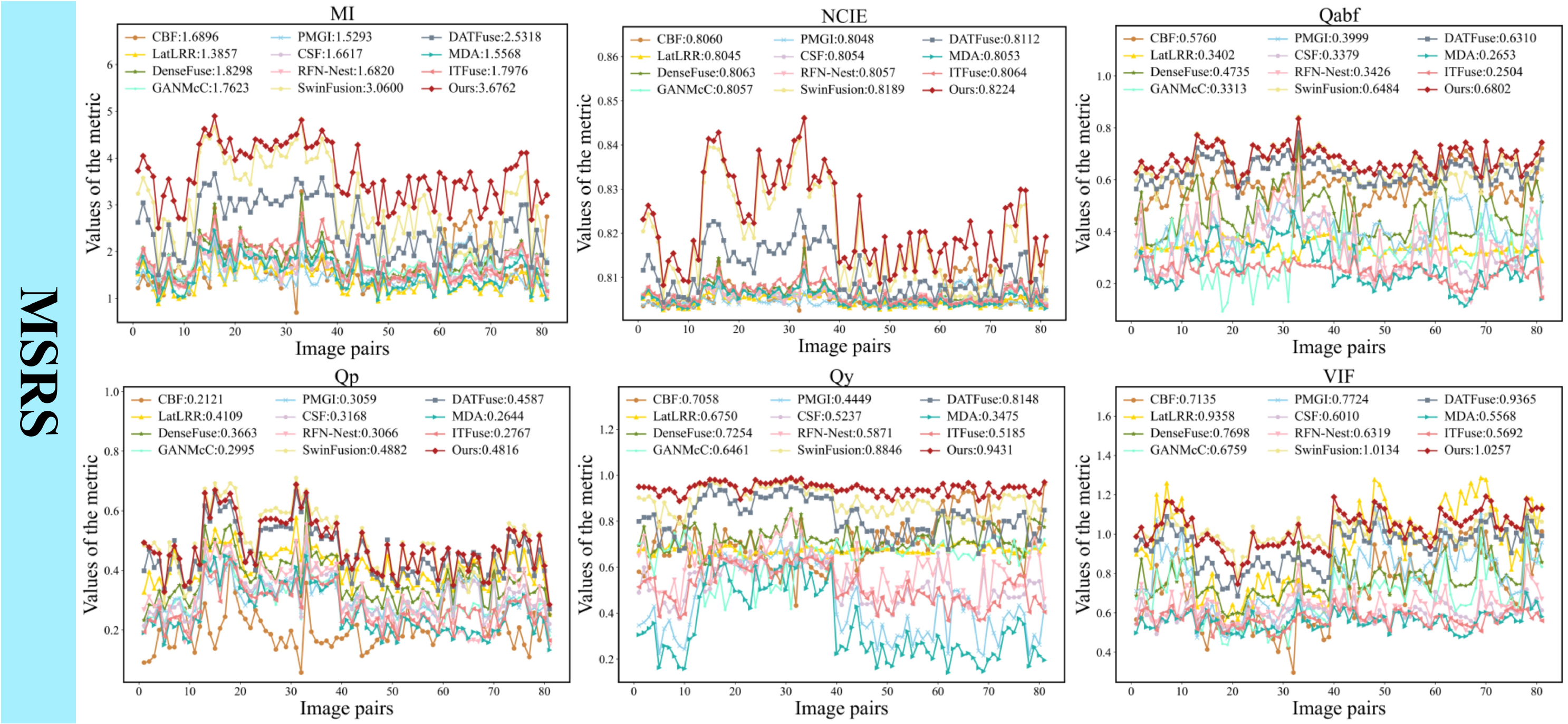}
\caption{Quantitative comparison of FSATFusion and eleven comparison methods on MSRS dataset. Our method is represented by the red line. Mean values of each method are shown in each legend. For the six metrics used for comparison, a higher value indicates superior performance.}
\label{Quantitative_MSRS}
\end{figure*}

\subsection{Comparative Experiments on TNO and MSRS datasets}
\label{sec:4c}

\textbf{Qualitative experiments on TNO dataset:}
In the qualitative comparative analysis on the TNO dataset, we presented the fusion results from various methods on four distinct pairs of source images. To facilitate visual comparisons, we have highlighted specific regions using red and green boxes and provided enlarged views of these areas. As depicted in Fig. \ref{Qualitative_TNO}, all eleven comparative methods successfully fused detailed information from visible images and significant target information from infrared images, producing fusion results with complementary information. However, each method exhibited particular shortcomings. The CBF method demonstrated a limited capacity for extracting infrared information, resulting in unsatisfactory visual quality in prominent target areas, as exemplified by the red box in the first column. Moreover, CBF often introduced unacceptable distortion and noise, as observed in the fused images of the second and fourth columns. LatLRR achieved the highest overall brightness, effectively capturing both infrared and visible details. Nevertheless, this heightened brightness also generated considerable noise, detracting from the visual quality. Additionally, due to the contrasting brightness levels between the foreground and background of the source images, some information was lost like the tree highlighted in the red box of the third column. DenseFuse, PMGI, CSF, RFN-Nest, GANMcC, MDA, and ITFuse similarly displayed this flaw. DenseFuse, PMGI, MDA and ITFuse struggled with extracting salient features from infrared images, leading to insufficient saliency and indistinct edges, as seen in the green box of the first column. CSF and MDA fusion results contained substantial noise, causing blurred edges and diminished visual quality, particularly noticeable in the fourth column. RFN-Nest and ITFuse limited ability to extract complementary features resulted in blurry fusion images, with the houses in the red box of the second column becoming indistinct. GANMcC fusion images were overly dark, forfeiting detailed information, such as the tiles in the first column. SwinFusion and DATFuse produced fusion images with ample complementary features and commendable visual quality, though SwinFusion was marred by noise, visible in the boxed area of the fourth column. DATFuse lacked emphasis on salient targets, as seen in the red box of the first column and the second column's fusion results. Collectively, FSATFusion excelled at extracting complementary information from source images, delivering fusion results with appropriate brightness, crisp edges, and superior visual quality.

\begin{figure*}[!t]
\centering
\includegraphics[width=\textwidth]{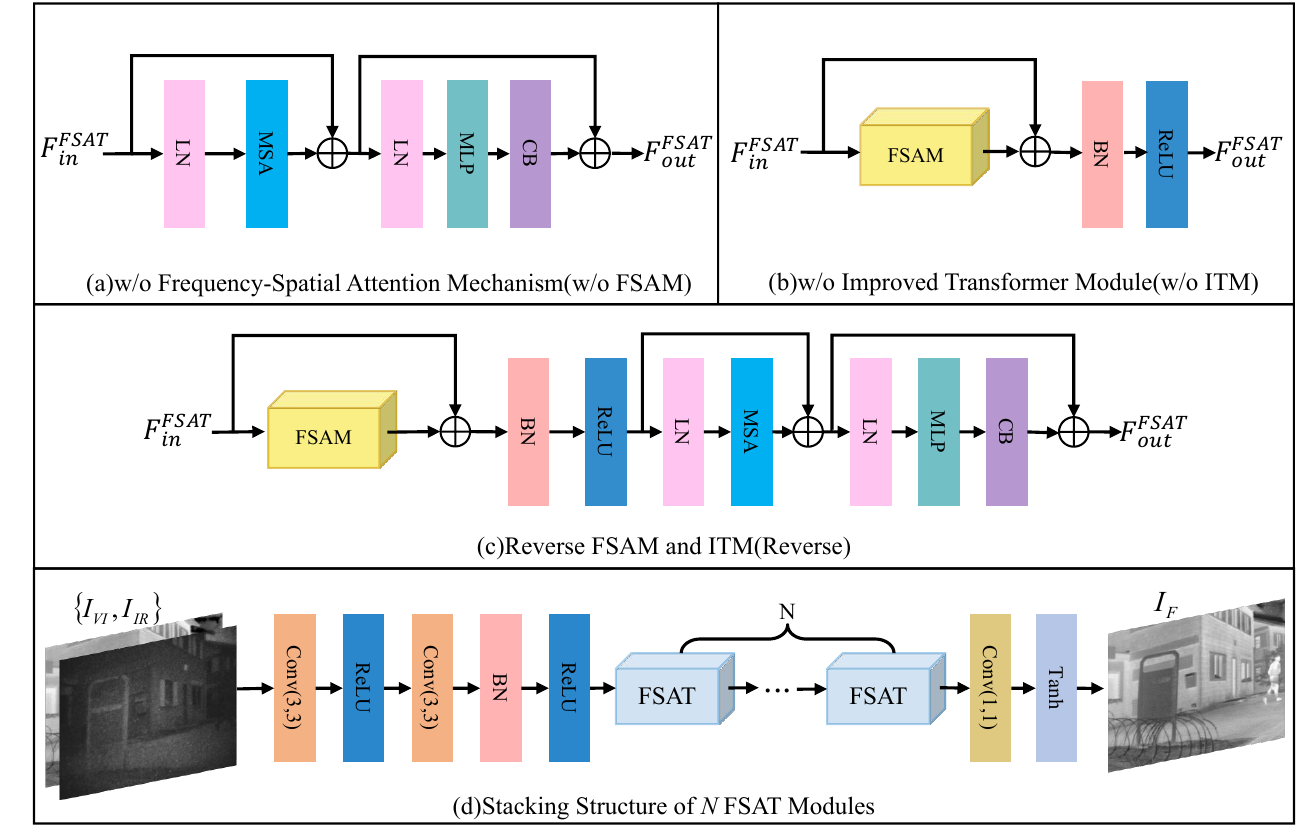}
\caption{Structure of network ablation experiment. We conducted structural analysis of FSAT using three structures: (a), (b), and (c). Meanwhile, we analyzed the stacking quantity of FSAT using the structure presented in (d).}
\label{Structure_ablation}
\end{figure*}

\begin{figure*}[!t]
\centering
\includegraphics[width=\textwidth]{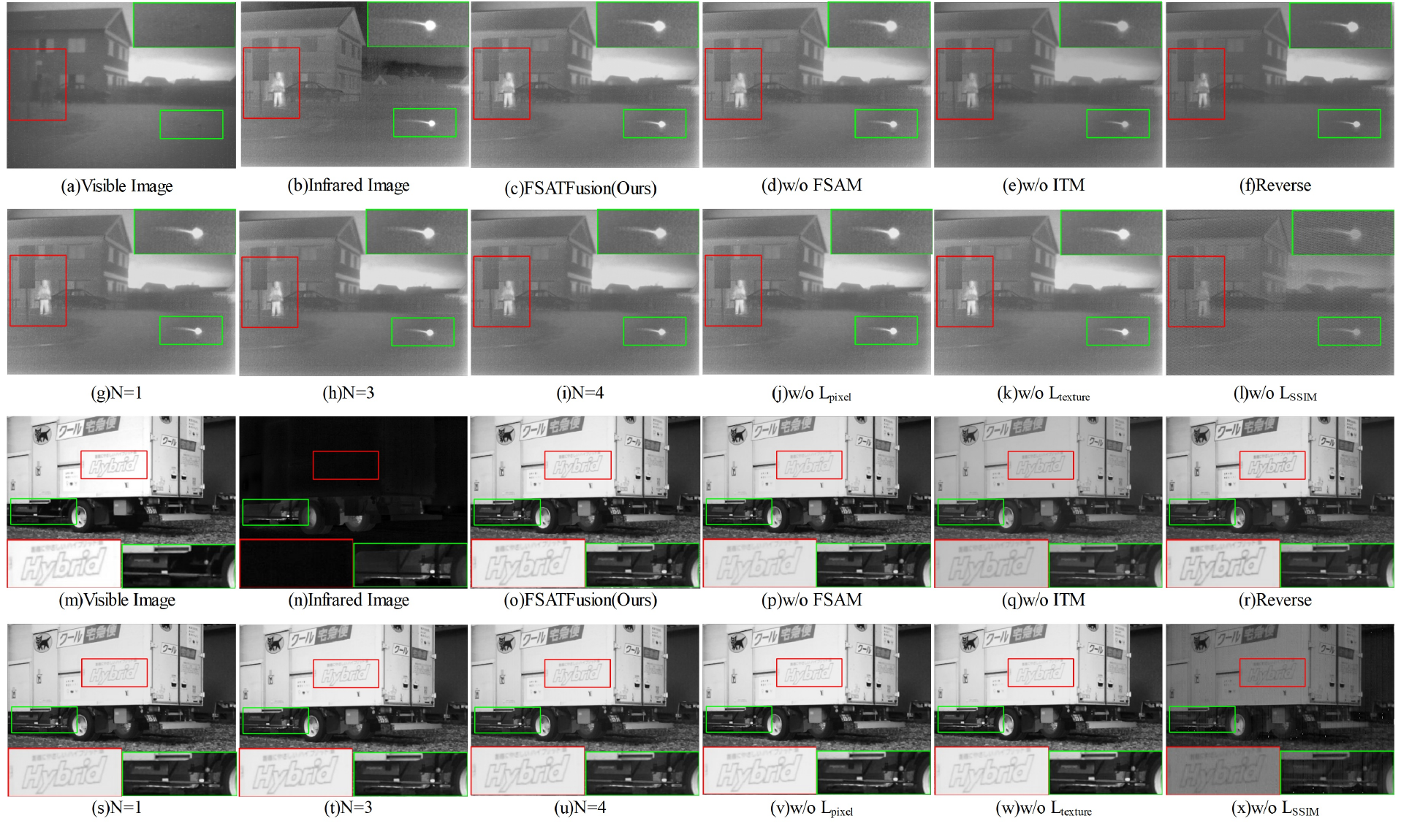}
\caption{Qualitative comparison of FSATFusion ablation experiment. For clearer comparison, we highlighted two important parts in each figure, marked with red and green boxes respectively.}
\label{Qualitative_ablation}
\end{figure*}

\begin{table*}[!t]
\centering
\caption{Quantitative comparison of ablation experiments. Red represents the best result. $\uparrow$ indicates that higher metrics reflect better fusion performance.}
\label{Quantitative_ablation}
\resizebox{0.8\textwidth}{!}{
\begin{tabular}{lccccccccc}
\toprule
 & & \textbf{MI$\uparrow$} & \textbf{NCIE$\uparrow$} & \textbf{$Q_{abf}\uparrow$} & \textbf{Q$_{P}$$\uparrow$} & \textbf{Q$_{y}$$\uparrow$} & \textbf{VIF$\uparrow$} \\
\midrule
\multirow{3}{*}{\textbf{Structure of FSAT}} & \textbf{w/o FSAM} & 3.7950 & 0.8192 &  0.5676 & 0.4444 & 0.9432 & 0.8598 \\
 & \textbf{w/o ITM} & 3.2461 & 0.8144 & 0.5176 & 0.4193 & 0.8650 & 0.7242 \\
 & \textbf{Reverse} & 3.2587 & 0.8149 & 0.5278 & 0.4244 & 0.8697 & 0.7513 \\
\midrule
\multirow{3}{*}{\textbf{Number of FSAT}} & \textbf{N=1} & 3.7621 & 0.8191 & 0.5784 & 0.4483 & 0.9508 & 0.8903 \\
 & \textbf{N=3} & 3.6947 & 0.8185 & 0.5740 & 0.4467 & 0.9471 & 0.8834 \\
 & \textbf{N=4} & 3.3093 & 0.8150 & 0.5561 & 0.4328 & 0.9223 & 0.8374 \\
\midrule
\multirow{3}{*}{\textbf{Loss Function}} & \textbf{w/o $L_{pixel}$} & 3.7770 & 0.8195 & 0.5747 & 0.4441 & 0.9449 & 0.8890 \\
 & \textbf{w/o $L_{texture}$} & 3.6090 & 0.8177 & 0.5680 & 0.4417 & 0.9413 & 0.8615 \\
 & \textbf{w/o $L_{SSIM}$} & 0.9552 & 0.8034 & 0.1946 & 0.1116 & 0.3397 & 0.2456 \\
\midrule
 & \textbf{FSATFusion (N=2)} & \textcolor{red}{3.9801} & \textcolor{red}{0.8214} & \textcolor{red}{0.5812} & \textcolor{red}{0.4489} & \textcolor{red}{0.9558} & \textcolor{red}{0.9033} \\
\bottomrule
\end{tabular}}
\end{table*}

\textbf{Quantitative experiments on TNO dataset:} In the quantitative comparative analysis on TNO dataset, the evaluation metrics comparing our method against eleven others are presented in Fig. \ref{Quantitative_TNO}. Our method consistently ranks first, demonstrating superior performance in preserving source image information (\(MI\)), maintaining image features such as gradients and edges (\(Q_{P}\)), showing strong nonlinear correlation with source images (\(NCIE\)), providing rich edge information (\(Q_{abf}\)), achieving good structural similarity (\(Q_{y}\)) and ensuring high visual perception quality (\(VIF\)).

\textbf{Qualitative experiments on MSRS dataset:} For the MSRS dataset, Fig. \ref{Qualitative_MSRS} presents the fusion results of various methods on four pairs of source images. Similar to the TNO dataset findings, all eleven methods generated fusion images with complementary information, yet each presented distinct drawbacks. CBF exhibited persistent distortion, leading to significant information loss, as evidenced in the first column. LatLRR's fusion results, while appropriately bright, were accompanied by noise that impacted visual quality, such as the road sign in the fourth column. DenseFuse, MDA and, ITFuse's limited ability to extract infrared features resulted in an inability to effectively highlight prominent targets, such as the person in the red box of the first column. PMGI failed to retain some visible light information during fusion, evident in the red and green boxes of the second and fourth columns. CSF's fusion results were generally dark, lacking in prominent target effects and losing detailed information, as highlighted in the green boxes of the second and third columns. RFN-Nest, GANMcC and, MDA did not effectively emphasize prominent infrared information, leading to blurry fusion images with indistinct edges, as observed in the red boxes of the first and third columns. SwinFusion and ITFuse results were dim, losing edge information (green box in the second column), while DATFuse's outcomes were noisy, causing blurred edges (green box in the fourth column). Overall, FSATFusion excelled at preserving details and edge information in visible light images (green box in the fourth column) and extracting salient features from infrared images, effectively accentuating them in the fusion results (red boxes in the first and second columns).

\textbf{Quantitative experiments on MSRS dataset:} As shown in Fig. \ref{Quantitative_MSRS}, the quantitative evaluation metrics comparing our method with eleven others on MSRS dataset \(Q_{P}\) is slightly suboptimal, its gap from the best-performing SwinFusion is lower. The other five metrics indicate leading position of FSATFusion. Note that VIF values typically range from \([0,1]\). In this study, we calculated the VIF of the fusion results independently for each source image and then summed the results, which explains why the obtained values may exceed 1.

Overall, the qualitative and quantitative comparison experiments on both TNO and MSRS datasets validate that the proposed FSATFusion effectively extracts and integrates complementary information from source images, achieving fusion results with super visual effects. This underscores the superiority fusion performance of FSATFsuion.

\begin{figure}[!t]
\centering
\includegraphics[width=\columnwidth]{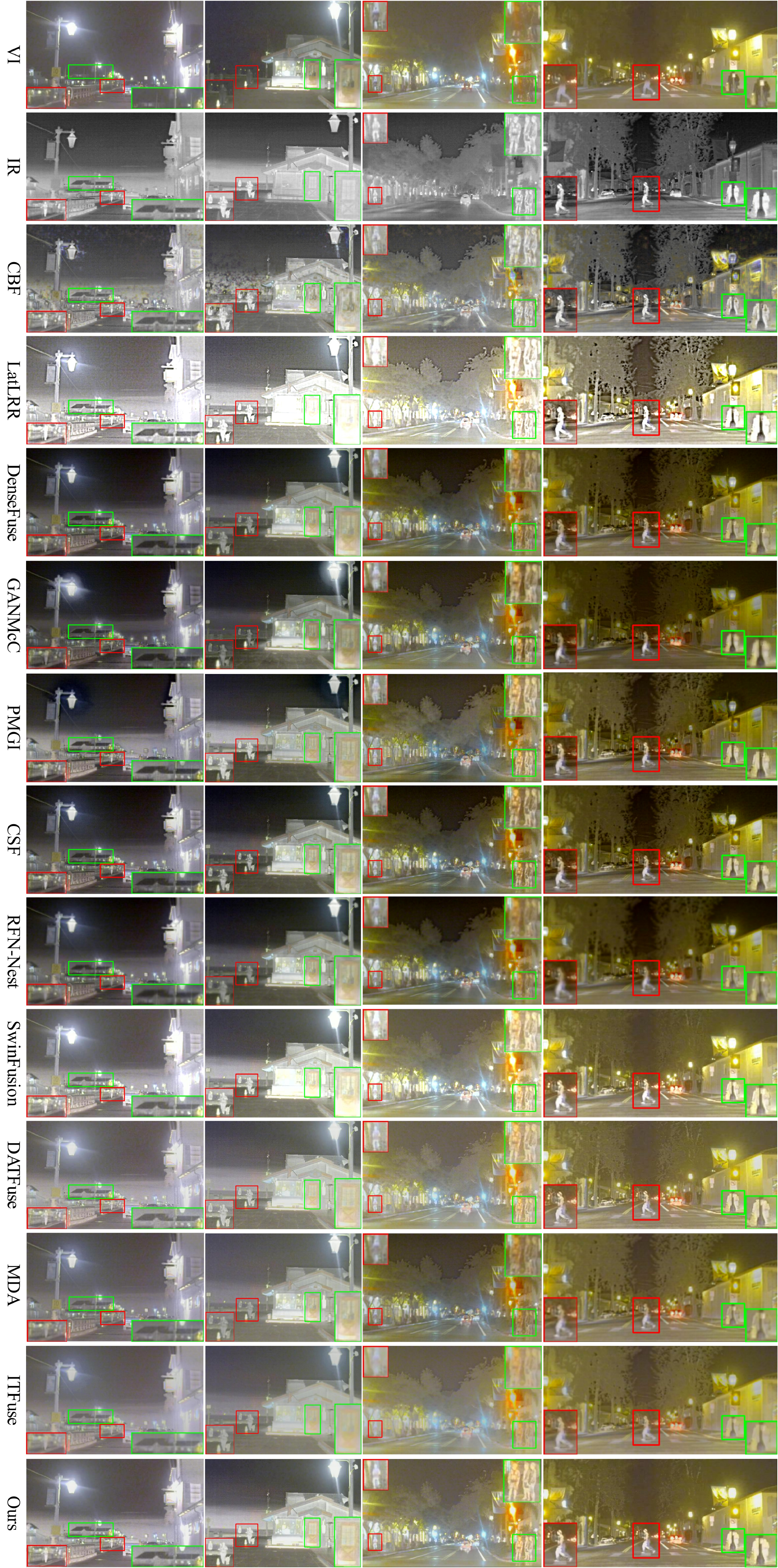}
\caption{Qualitative comparison of FSATFusion and eleven comparison methods on RoadScene dataset. For clearer comparison, we highlighted and enlarged two important parts in each figure, marked with red and green boxes respectively.}
\label{Qualitative_RoadScene}
\end{figure}

\begin{figure}[!t]
\centering
\includegraphics[width=\columnwidth]{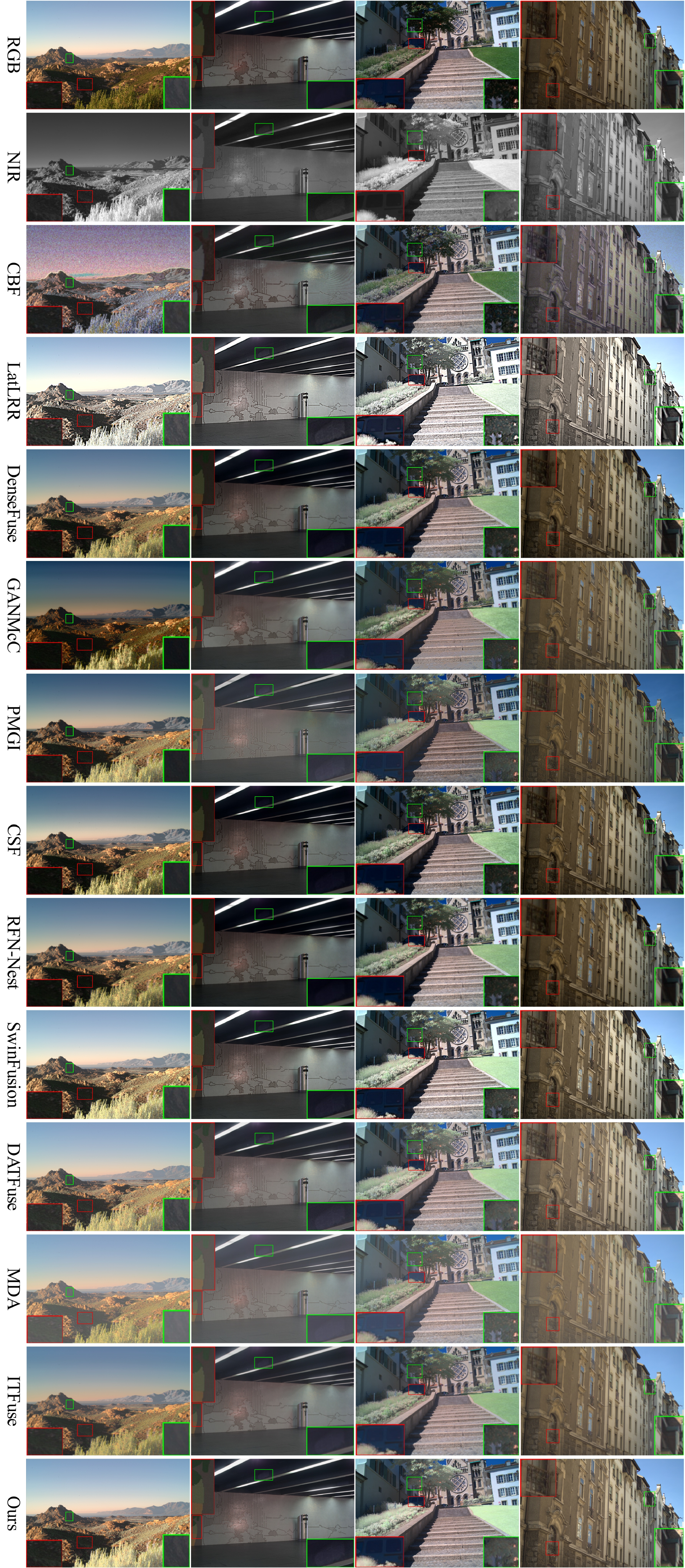}
\caption{Qualitative comparison of FSATFusion and eleven comparison methods on RGB-NIR dataset. For clearer comparison, we highlighted and enlarged two important parts in each figure, marked with red and green boxes respectively.}
\label{Qualitative_RGBNIR}
\end{figure}

\subsection{Ablation Study}
\label{sec:4d}
To rigorously demonstrate the efficacy and design rationale of the FSATFusion network, a comprehensive suite of ablation studies was executed. The ITM and FSAM, integral components of our FSAT module, are essential in extracting global features and directing the network's focus onto critical areas of the feature map. These elements are integral to the entire FSATFusion network. Specifically, initial ablation experiments involved the removing and repositioning the ITM and FSAM within the FSAT module. Subsequently, we performed stacking layer experiments on FSAT to ascertain the network structure's rationality. To demonstrate the necessity of each loss function, ablation experiments were conducted on the three loss components utilized within the network. The outcomes were analyzed both qualitatively and quantitatively, with the network configurations for each ablation experiment illustrated in Fig. \ref{Structure_ablation} and the qualitative comparison outcomes presented in Fig. \ref{Qualitative_ablation}.

\begin{figure*}[!t]
\centering
\includegraphics[width=\textwidth]{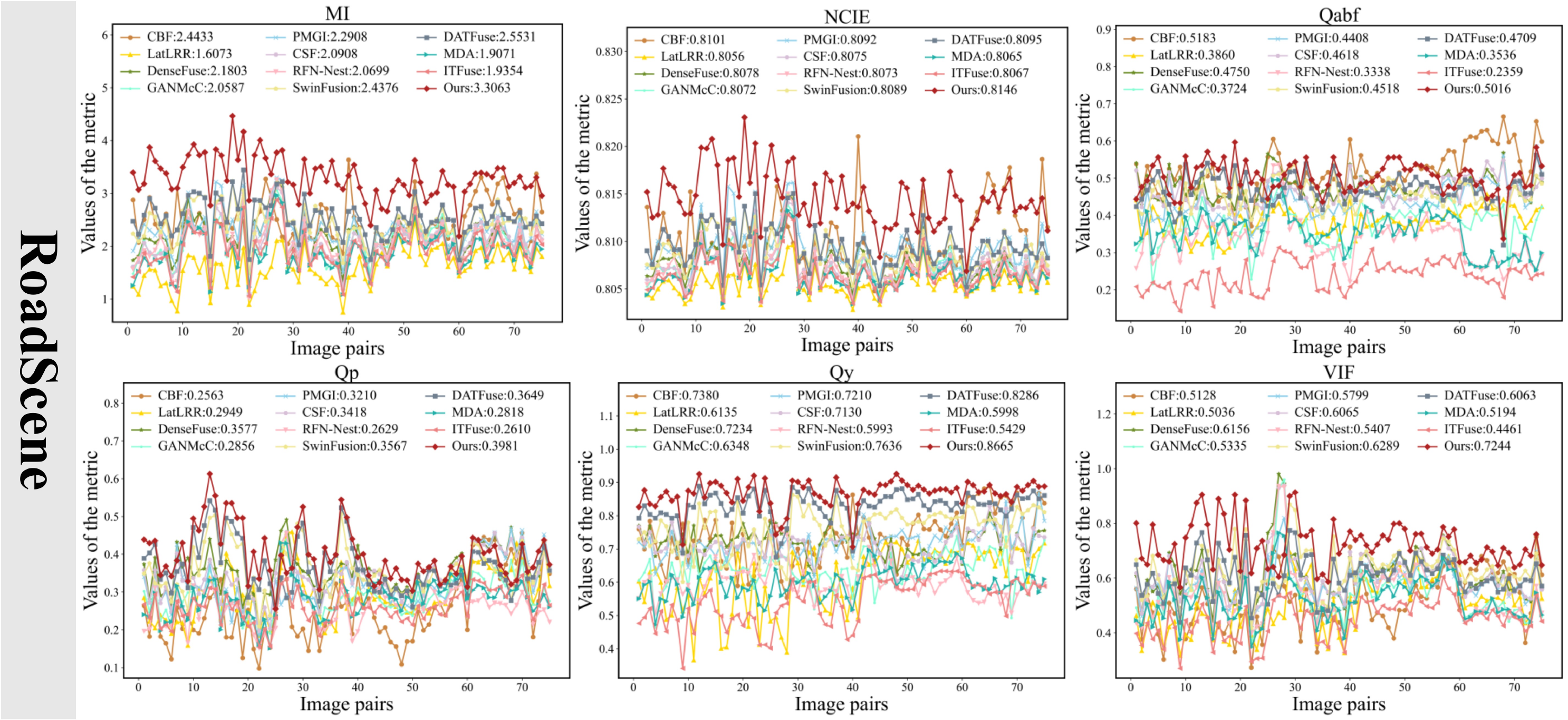}
\caption{Quantitative comparison of FSATFusion and eleven comparison methods on RoadScene dataset. Our method is represented by the red line. Mean values of each method are shown in each legend. For the six metrics used for comparison, a higher value indicates superior performance.}
\label{Quantitative_RoadScene}
\end{figure*}

\textbf{Structural Analysis of FSAT:} Ecluding FSAM (w/o FSAM) led to a significant reduction in the network's capacity to extract salient infrared targets and visible light details, as highlighted by the green box in the first set of results and the red box in the second set of results in Fig. \ref{Qualitative_ablation}. Removing ITM from FSAT (w/o ITM) resulted in the network's inability to effectively amalgamate complementary information from the dual source images, evident in the diminished visibility of significant target information within the red box of the first set's results and the green box of the second set's results. Furthermore, reversing the order of FSAM and ITM (Reverse) caused an overall reduction in brightness in the first set, culminating in the loss of detailed features. This suggests that the attention mechanism may inadvertently suppress certain feature components map, leading to their underemphasis in downstream processing and subsequent feature degradation.

\textbf{Analysis of FSAT Quantity:} Paralleling the behavior of traditional CNNs, the performance of our FSATFusion does not invariably escalate with an increasing number of FSAT modules connected in series. To validate the rationality of our FSATFusion structure, experiments were conducted to vary the quantity of FSAT modules dedicated to global feature extraction and focal feature emphasis. Configurations with N=1, N=3, and N=4 were tested and compared against our original setting of FSATFusion (N=2). Qualitative assessments indicated that the fusion outcomes for N=1 and N=4 were generally darker. Although N=3 exhibited similar overall brightness, and the second set's results demonstrated reduced edge contrast with N=3. Collectively, the configuration with N=2 yielded the most favorable visual outcome, thereby confirming the rationality of our structural design.

\textbf{Analysis of Loss Function:} We conducted qualitative and quantitative experiments on pixel loss \(\mathcal{L}_{pixel}\), texture loss \(\mathcal{L}_{texture}\), and SSIM loss \(\mathcal{L}_{SSIM}\) to demonstrate their necessity and effectiveness. Without \(\mathcal{L}_{pixel}\) the edge detail contrast decreased (red box in the second group's results in Fig. \ref{Qualitative_ablation}). Ablating \(\mathcal{L}_{texture}\) resulted in lower quality of edge and detail information from visible images (red box in the second group's results in Fig. \ref{Qualitative_ablation}). Without \(\mathcal{L}_{SSIM}\), the network failed to highlight infrared target information effectively, and noticeable stripe distortion appeared (red box in the first and second group's results in Fig. \ref{Qualitative_ablation}).

Tab. \ref{Quantitative_ablation} presents a comparative experiment of quantitative evaluation metrics across variation network structures, with the best-performing metrics highlighted in red. The lack of attention mechanism led to decreased \(VIF\), reducing visual quality. The substantial decrease in \(MI\), \(Q_{abf}\), \(Q_{P}\), and \(VIF\) after ITM ablation highlights ITM's importance in preserving source image information and image features, affecting the visual quality of fusion results. Altering the relative positions of ITM and FSAM showed that performing attention computation first negatively impacted source image information and feature retention, as indicated by the reduced \(Q_{y}\). Experiments with varying numbers of FSAT modules indicated that a single FSAT module was insufficient for the adequate extraction of advanced features. An increase in Transformer layers also resulted in the attenuation or loss of source image information and features during the layer-by-layer transmission. The ablation study of loss function, the removal of either the pixel loss or texture loss function decreased evaluation metrics and impaired the visual effect. Training devoid of SSIM loss guidance led to severe metric deterioration, highlighting the essential function of the SSIM loss function in FSATFusion training.

In conclusion, the qualitative comparison experiments and quantitative comparisons regarding the impacts of structural modifications are congruent. Even minor structural alterations can lead to significant variations in quantitative evaluation metrics. Summarizing the findings, any modification to the network structure or loss function composition of FSATFusion results in degradation of fusion performance to varying extents and in different aspects.

\begin{figure*}[!t]
\centering
\includegraphics[width=\textwidth]{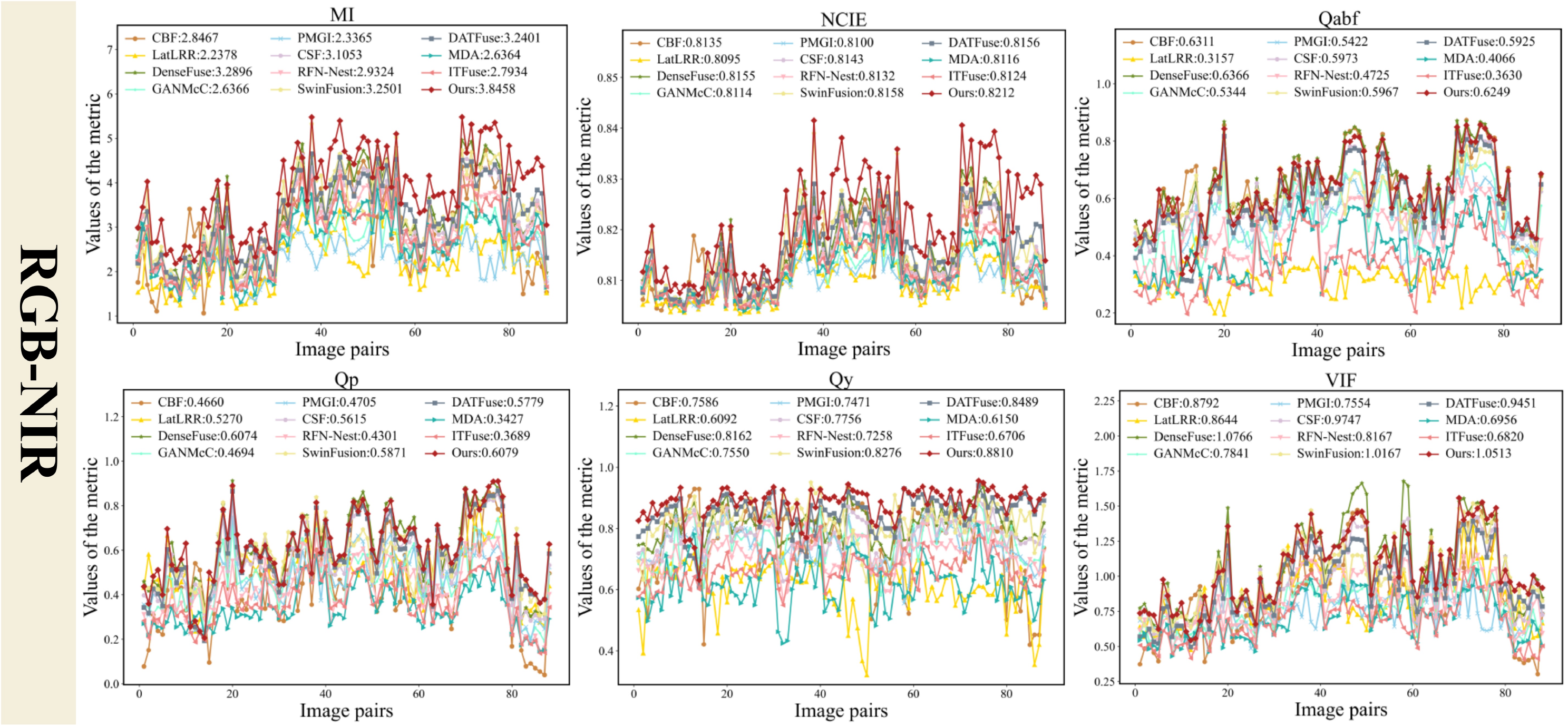}
\caption{Quantitative comparison of FSATFusion and eleven comparison methods on RGB-NIR dataset. Our method is represented by the red line. Mean values of each method are shown in each legend. For the six metrics used for comparison, a higher value indicates superior performance.}
\label{Quantitative_RGBNIR}
\end{figure*}

\subsection{Generalization Experiment}
\label{sec:4e}

To demonstrate the generalization ability of FSATFusion, we extended its application scenes: infrared-RGB image fusion and NIR-RGB image fusion. Qualitative and quantitative comparative experiments were conducted using the RoadScene and RGB-NIR datasets, respectively, employing the model without any alterations.

\subsubsection{\textbf{Infrared and RGB Image Fusion}} 
For infrared and RGB image fusion, we first convert the three-channel visible image into a YUV image and extract the brightness channel component \(I_{RGB}^{Y}\). This component is then fused with the infrared image \(I_{IR}\) to obtain \(I_{Y}^{F}\). Finally, \(I_{F}^{Y}\) is recombined with the U, V components \(I_{RGB}^{U}\), \(I_{RGB}^{V}\)  of the initial RGB image and converted back into an RGB image to obtain the final fusion result \(I_{F}\).

\begin{table*}[!t]
\centering
\caption{Comparison of running time(s) between FSATFusion and eleven comparison methods on four datasets. Average represents the average running time of four datasets. Red represents the best outcome.}
\label{time}
\resizebox{\textwidth}{!}{
\begin{tabular}{lccccccccccccc}
\hline
 & \textbf{} & \textbf{CBF} & \textbf{LatLRR} & \textbf{DenseFuse} & \textbf{PMGI} & \textbf{CSF} & \textbf{RFN-Nest} & \textbf{GANMcC} & \textbf{SwinFusion} & \textbf{DATFuse} & \textbf{MDA} & \textbf{ITFuse} & \textbf{Ours} \\ \hline
\multirow{4}{*}{\textbf{}} & \textbf{TNO} & 5.575 & 87.462 & 0.177 & 0.156 & 4.640 & 0.118 & 0.479 & 2.400 & 0.066 & 2.599 & 0.061 & \textcolor{red}{0.052} \\ 
 & \textbf{MSRS} & 5.097 & 82.666 & 0.182 & \textcolor{red}{0.043} & 3.041 & 0.087 & 0.115 & 1.588 & 0.054 & 4.705 & 0.051 & 0.046 \\ 
 & \textbf{RoadScene} & 9.226 & 139.814 & 0.373 & 0.136 & 2.273 & 0.049 & 0.603 & 0.811 & 0.037 & 3.513 & 0.062 & \textcolor{red}{0.034} \\ 
 & \textbf{RGB-NIR} & 34.130 & 502.830 & 0.201 & \textcolor{red}{0.087} & 6.070 & 0.181 & 0.257 & 3.709 & 0.119 & 10.03 & 0.126 & 0.096 \\ \hline
\textbf{}&\textbf{Average} & 13.507 & 203.193 & 0.233 & 0.105 & 4.006 & 0.109 & 0.363 & 2.127 & 0.069 & 5.212 & 0.075 & \textcolor{red}{0.057} \\ \hline
\end{tabular}}
\end{table*}

\textbf{Qualitative comparative experiments on RoadScene dataset:} As shown in Fig. \ref{Qualitative_RoadScene}, we selected four pairs of source images from RoadScene dataset and applied eleven comparative methods to obtain fusion results. Although all eleven methods compared to FSATFusion produced fusion results with complementary information from visible and infrared images, they had notable disadvantages. CBF results showed partial distortion and low color fidelity (fusion result of the second column). LatLRR and SwinFusion exhibited excessive brightness enhancement, losing detail information (the gate details in the second column). DenseFuse, DATFuse, MDA, and ITFuse had limited ability to preserve infrared salient features, resulting in low visibility and saliency of these targets (the person in the red box of the first case). PMGI's results showed unnatural brightness transitions around street lamps (first and third columns). CSF and ITFuse encountered difficulties in extracting edge information from infrared images, leading to low-contrast edges, such as those of the house in the green box in the first column. RFN-Nest, GANMcC and, MDA delivered blurred results with inadequate target saliency. In conclusion, FSATFusion proficiently extracted complementary information and preserved visible color details, ensuring optimal visual quality.

\textbf{Quantitative comparative experiments on RoadScene dataset:} Quantitative comparative experiments conducted on the RoadScene dataset reveal that while FSATFusion did not secure the highest scores in the \(Q_{abf}\) metric, as illustrated in Fig. \ref{Quantitative_RoadScene}, it surpassed the majority of the comparative methods in overall performance. FSATFusion attained the best results in five additional evaluation metrics, showcasing considerable superiority over other techniques.

\subsubsection{\textbf{Near-Infrared and RGB image fusion}}
To further investigate the generalization performance, we applied it to fusing near-infrared images with RGB images, using the same approach as the infrared and visible image fusion task.

\textbf{Qualitative comparative experiments on RGB-NIR dataset:} As shown in Fig. \ref{Qualitative_RGBNIR}, eleven different fusion methods were applied to four pairs of source images from RGB-NIR dataset. Compared to FSATFusion, other eleven methods had notable drawbacks. The CBF method introduced common noise and compromised color fidelity, particularly noticeable in the sky of the first column. LatLRR maintained edge details in darker regions but suffered from low color fidelity due to excessive brightness enhancement, which was evident in the overall color fidelity of the first and third columns. DenseFuse and CSF were unable to capture near-infrared detail edge features, leading to a loss of detail information in low-light areas, as indicated by the green box in the second column. PMGI and GANMcC produced insufficient brightness, resulting in a loss of visible light details in low-light areas, as seen in the green box of the first column. RFN-Nest, MDA, and ITFuse failed to effectively utilize high-frequency information, causing blurred edges, exemplified by the house edges in the green box of the first and fourth columns. Transformer-based methods, while effective in extracting detail information, had specific deficiencies; SwinFusion did not sufficiently highlight details in low-light areas (red box in the third column), and DATFuse exhibited slightly insufficient color fidelity (red box in the second column). In summary, FSATFusion adeptly utilized near-infrared information to complement missing details caused by shadows in visible images, yielding fused images that are rich in detail and possess excellent visual effects.

\textbf{Quantitative comparative experiments on RGB-NIR dataset:} Fig. \ref{Quantitative_RGBNIR} compares the quantitative evaluation metrics of FSATFusion and eleven other methods for near-infrared and visible image fusion. FSATFusion achieved the best performance in  \(MI\), \(NCIE\), \(Q_{P}\), and \(Q_{y}\). Although  \(Q_{abf}\), and \(VIF\) did not achieve maximum values, their performance is superior to most comparison methods.

The comprehensive qualitative and quantitative comparisons demonstrate that FSATFusion is effective in fusing infrared images with RGB images and near-infrared images with RGB images. The fusion outcomes not only display pleasing visual effects but also excel in information extraction, feature preservation, structural similarity, and visual fidelity. These experiments substantiate the strong generalization capability of FSATFusion.

\begin{figure}[!t]
\centering
\includegraphics[width=\columnwidth]{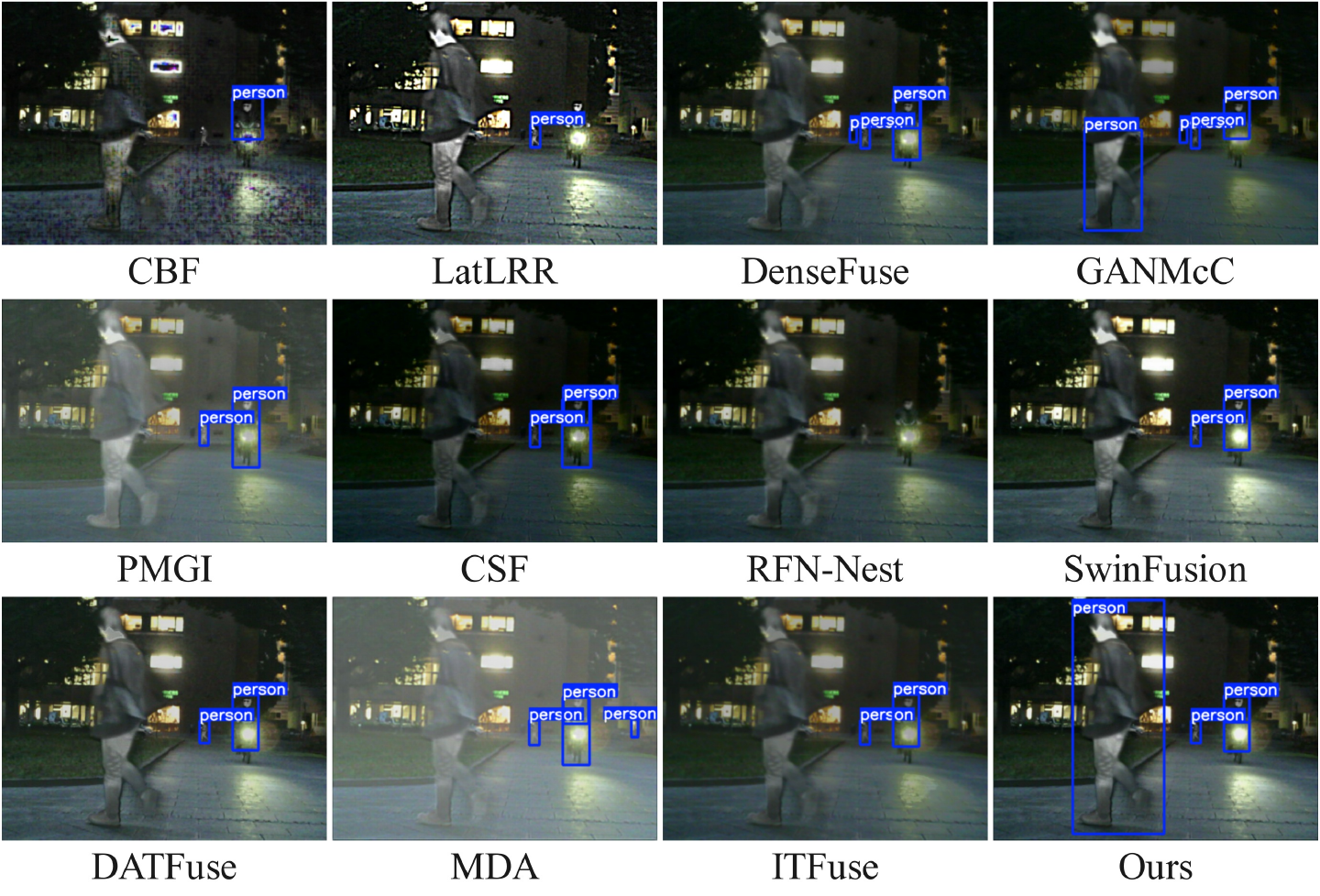}
\caption{Detection performance of FSATFusion and eleven comparison methods on labeled image from MSRS dataset.}
\label{Qualitative_detection}
\end{figure}

\subsection{Operational Efficiency Analysis}
\label{sec:4f}

As shown in Tab. \ref{time}, a comparison of the execution speed of FSATFusion with eleven alternative methods was conducted across four distinct datasets. FSATFusion exhibited the most rapid execution on both the TNO and RoadScene datasets. While its efficiency on the MSRS and RGB-NIR datasets was marginally less than that of the PMGI fast fusion framework, the average runtime of FSATFusion across all four datasets was considerably lower than PMGI, achieving the best performance among all the methods under comparison. These findings underscore the effectiveness of FSATFusion's compact network architecture in enhancing the operational efficiency typical of Transformer-based models. FSATFusion adeptly balances high operational efficiency with superior fusion performance.

\subsection{Detection Performance}
\label{sec:4g}

%在本节中,我们旨在证明我们的方法应用于下游任务的优越性。我们将MSRS数据集中80对含检测标签的图像使用不同的融合算法进行融合，并将融合结果输入到YOLOv5模型上进行目标检测。如图X所示，图片场景中有三个行人，其中一个行人是小而原的，还有一个行人因为快速移动而产生了拖影.除了我们的FSATFusion得到的融合结果能够得到满意的检测结果外，其他9种对比方法都存在着一些缺点：DenseFuse和GANMcC的融合图像产生了错误的探测结果，RFN-Nest的融合结果在目标检测中无法得到检测结果，其他的6种对比方法得到的融合结果在目标检测中都存在缺少目标的情况。
%为了测量不同融合方法应用于目标检测任务的平均精度，我们还进行了定量实验。如表X所示，我们展示了Iou为不同阈值时的mAP，从结果可以看出，我们的FSATFusion在不同的IoU阈值上都具有良好的mAP值，这表明我们的融合结果相比于其他融合方法可以更有效地应用于下游任务。

\begin{table}[!t]
\centering
\caption{Quantitative comparison of detection performance on MSRS dataset. RED represent the best result.}
\label{Quantitative_detection}
\resizebox{\columnwidth}{!}{
\begin{tabular}{cccc}
\hline
\textbf{} & \textbf{mAP@0.65} &  \textbf{mAP@0.85} & \textbf{mAP@[0.5,0.95]} \\
\hline
CBF\cite{shreyamsha2015image}        & 0.751 &  0.414 & 0.571 \\
LatLRR\cite{li2018infrared}     & 0.765 &  0.438 & 0.600 \\
DenseFuse\cite{li2018densefuse}  & 0.835 &  0.486 & 0.639 \\
PMGI\cite{zhang2020rethinking}       & 0.824 &  0.445 & 0.613 \\
CSF\cite{xu2021classification}        & 0.827 &  0.508 & 0.640 \\
RFN-Nest\cite{li2021rfn}   & 0.775 &  0.476 & 0.593 \\
GANMcC\cite{ma2020ganmcc}     & 0.842 & 0.534 & 0.656 \\
SwinFusion\cite{ma2022swinfusion} & 0.830 & 0.477 & 0.637 \\
DATFuse\cite{tang2023datfuse}    & 0.843 & 0.509 & 0.658 \\
MDA\cite{yang2024mda}   & 0.792 & 0.382 & 0.590 \\
ITFuse\cite{tang2024itfuse}   & 0.842 & 0.494 & 0.644 \\
Ours       & \textcolor{red}{0.844} &  \textcolor{red}{0.549} & \textcolor{red}{0.658} \\
\hline
\end{tabular}}
\end{table}

To demonstrate the superiority of FSATFusion when applied to downstream tasks. We utilized various fusion algorithms on 80 pairs of labeled images from the MSRS dataset and subsequently input the fused results into the YOLOv5 \cite{yolov5} model for object detection. As depicted in Fig. \ref{Qualitative_detection}, the scene includes three pedestrians: one is small and distant, while another exhibits motion blur due to rapid movement. Among the compared methods, only our FSATFusion provided satisfactory detection outcomes. The other eleven methods showed deficiencies: DenseFuse, GANMcC and MDA produced incorrect detection results, RFN-Nest failed to generate any detection results, and the remaining seven methods omitted targets in their fused outputs.

To evaluate the average precision of different fusion algorithms for object detection, quantitative experiments were conducted. As shown in Tab. \ref{Quantitative_detection}, we present the mean average precision (mAP) for various intersection over union (IoU) thresholds. The results indicate that FSATFusion consistently attains high mAP values across different IoU thresholds, underscoring its superiority in downstream tasks relative to other fusion methods.

\section{Conclusion}
\label{sec:conclusion}
This paper presents FSATFusion, a novel infrared and visible image fusion (IVIF) network based on a frequency-spatial attention Transformer. FSATFusion enhances the global feature extraction capabilities of standard Transformer architectures through its Improved Transformer Module (ITM) and leverages a Frequency-Spatial Attention Mechanism (FSAM) to selectively extract critical features while preserving multi-domain information. Extensive comparative experiments on four benchmark datasets against eleven advanced traditional and deep learning methods. Results demonstrate that FSATFusion achieves superior performance in both qualitative and quantitative evaluations, highlighting its excellent fusion capability, strong generalization ability, and high computational efficiency. Moreover, Downstream object detection experiments further validate its effectiveness in enhancing visual task performance, highlighting its practical utility beyond direct fusion. Future work will explore the application of FSATFusion to the fusion of unregistered images, further broadening its applicability to other image fusion tasks such as multi-exposure and multi-focus fusion. Overall, we strive to develop a unified, efficient, and high-quality framework image fusion applications.

\bibliographystyle{IEEEtran}
\bibliography{reference}

\begin{thebibliography}{10}
\providecommand{\url}[1]{#1}
\csname url@samestyle\endcsname
\providecommand{\newblock}{\relax}
\providecommand{\bibinfo}[2]{#2}
\providecommand{\BIBentrySTDinterwordspacing}{\spaceskip=0pt\relax}
\providecommand{\BIBentryALTinterwordstretchfactor}{4}
\providecommand{\BIBentryALTinterwordspacing}{\spaceskip=\fontdimen2\font plus
\BIBentryALTinterwordstretchfactor\fontdimen3\font minus \fontdimen4\font\relax}
\providecommand{\BIBforeignlanguage}[2]{{%
\expandafter\ifx\csname l@#1\endcsname\relax
\typeout{** WARNING: IEEEtran.bst: No hyphenation pattern has been}%
\typeout{** loaded for the language `#1'. Using the pattern for}%
\typeout{** the default language instead.}%
\else
\language=\csname l@#1\endcsname
\fi
#2}}
\providecommand{\BIBdecl}{\relax}
\BIBdecl

\bibitem{hu2025datransnet}
C.~Hu, Y.~Huang, K.~Li, L.~Zhang, C.~Long, Y.~Zhu, T.~Pu, and Z.~Peng, ``Datransnet: dynamic attention transformer network for infrared small target detection,'' \emph{IEEE Geoscience and Remote Sensing Letters}, 2025.

\bibitem{xiao2024background}
M.~Xiao, Q.~Dai, Y.~Zhu, K.~Guo, H.~Wang, X.~Shu, J.~Yang, and Y.~Dai, ``Background semantics matter: Cross-task feature exchange network for clustered infrared small target detection with sky-annotated dataset,'' \emph{arXiv preprint arXiv:2407.20078}, 2024.

\bibitem{zhang2020object}
X.~Zhang, P.~Ye, H.~Leung, K.~Gong, and G.~Xiao, ``Object fusion tracking based on visible and infrared images: A comprehensive review,'' \emph{Information Fusion}, vol.~63, pp. 166--187, 2020.

\bibitem{xiao2020review}
Y.~Xiao, Z.~Tian, J.~Yu, Y.~Zhang, S.~Liu, S.~Du, and X.~Lan, ``A review of object detection based on deep learning,'' \emph{Multimedia Tools and Applications}, vol.~79, pp. 23\,729--23\,791, 2020.

\bibitem{wang2014fusion}
J.~Wang, J.~Peng, X.~Feng, G.~He, and J.~Fan, ``Fusion method for infrared and visible images by using non-negative sparse representation,'' \emph{Infrared Physics \& Technology}, vol.~67, pp. 477--489, 2014.

\bibitem{selvaraj2020infrared}
A.~Selvaraj and P.~Ganesan, ``Infrared and visible image fusion using multi-scale nsct and rolling-guidance filter,'' \emph{IET Image Processing}, vol.~14, no.~16, pp. 4210--4219, 2020.

\bibitem{li2018densefuse}
H.~Li and X.-J. Wu, ``Densefuse: A fusion approach to infrared and visible images,'' \emph{IEEE Transactions on Image Processing}, vol.~28, no.~5, pp. 2614--2623, 2018.

\bibitem{stfnet}
Q.~Liu, J.~Pi, P.~Gao, and D.~Yuan, ``Stfnet: Self-supervised transformer for infrared and visible image fusion,'' \emph{IEEE Transactions on Emerging Topics in Computational Intelligence}, vol.~8, no.~2, pp. 1513--1526, 2024.

\bibitem{wang2024uud}
X.~Wang, L.~Fang, J.~Zhao, Z.~Pan, H.~Li, and Y.~Li, ``Uud-fusion: An unsupervised universal image fusion approach via generative diffusion model,'' \emph{Computer Vision and Image Understanding}, vol. 249, p. 104218, 2024.

\bibitem{luo2024hbanet}
X.~Luo, J.~Zhang, L.~Wang, and D.~Niu, ``Hbanet: A hybrid boundary-aware attention network for infrared and visible image fusion,'' \emph{Computer Vision and Image Understanding}, vol. 249, p. 104161, 2024.

\bibitem{huang2024fusiondiff}
Z.~Huang, S.~Yang, J.~Wu, L.~Zhu, and J.~Liu, ``Fusiondiff: A unified image fusion network based on diffusion probabilistic models,'' \emph{Computer Vision and Image Understanding}, vol. 244, p. 104011, 2024.

\bibitem{zhang2025exploring}
T.~Zhang, Y.~Zhu, J.~Zhao, G.~Cui, and Y.~Zheng, ``Exploring state space model in wavelet domain: An infrared and visible image fusion network via wavelet transform and state space model,'' \emph{arXiv preprint arXiv:2503.18378}, 2025.

\bibitem{zhang2025daaf}
T.~Zhang, J.~Zhao, Y.~Zhu, G.~Cui, Y.~Jing, and Y.~Lyu, ``Daaf: Degradation-aware adaptive fusion framework for robust infrared and visible images fusion,'' \emph{arXiv preprint arXiv:2504.10871}, 2025.

\bibitem{liu2023sgfusion}
J.~Liu, R.~Dian, S.~Li, and H.~Liu, ``Sgfusion: A saliency guided deep-learning framework for pixel-level image fusion,'' \emph{Information Fusion}, vol.~91, pp. 205--214, 2023.

\bibitem{xu2020u2fusion}
H.~Xu, J.~Ma, J.~Jiang, X.~Guo, and H.~Ling, ``U2fusion: A unified unsupervised image fusion network,'' \emph{IEEE Transactions on Pattern Analysis and Machine Intelligence}, vol.~44, no.~1, pp. 502--518, 2020.

\bibitem{li2021rfn}
H.~Li, X.-J. Wu, and J.~Kittler, ``Rfn-nest: An end-to-end residual fusion network for infrared and visible images,'' \emph{Information Fusion}, vol.~73, pp. 72--86, 2021.

\bibitem{xu2022cufd}
H.~Xu, M.~Gong, X.~Tian, J.~Huang, and J.~Ma, ``Cufd: An encoder--decoder network for visible and infrared image fusion based on common and unique feature decomposition,'' \emph{Computer Vision and Image Understanding}, vol. 218, p. 103407, 2022.

\bibitem{ma2019fusiongan}
J.~Ma, W.~Yu, P.~Liang, C.~Li, and J.~Jiang, ``Fusiongan: A generative adversarial network for infrared and visible image fusion,'' \emph{Information fusion}, vol.~48, pp. 11--26, 2019.

\bibitem{apnet2024}
W.~Zhou, Y.~Zhu, J.~Lei, J.~Wan, and L.~Yu, ``Apnet: Adversarial learning assistance and perceived importance fusion network for all-day rgb-t salient object detection,'' \emph{IEEE Transactions on Emerging Topics in Computational Intelligence}, vol.~6, no.~4, pp. 957--968, 2022.

\bibitem{lgabl2024}
D.~Wang, Z.~Zheng, W.~Ding, and X.~Jia, ``Lgabl: Uhd multi-exposure image fusion via local and global aware bilateral learning,'' \emph{IEEE Transactions on Emerging Topics in Computational Intelligence}, vol.~8, no.~2, pp. 1362--1375, 2024.

\bibitem{dosovitskiy2020image}
A.~Dosovitskiy, L.~Beyer, A.~Kolesnikov, D.~Weissenborn, X.~Zhai, T.~Unterthiner, M.~Dehghani, M.~Minderer, G.~Heigold, S.~Gelly \emph{et~al.}, ``An image is worth 16x16 words: Transformers for image recognition at scale,'' \emph{arXiv preprint arXiv:2010.11929}, 2020.

\bibitem{ma2020ganmcc}
J.~Ma, H.~Zhang, Z.~Shao, P.~Liang, and H.~Xu, ``Ganmcc: A generative adversarial network with multiclassification constraints for infrared and visible image fusion,'' \emph{IEEE Transactions on Instrumentation and Measurement}, vol.~70, pp. 1--14, 2020.

\bibitem{tang2024itfuse}
W.~Tang, F.~He, and Y.~Liu, ``Itfuse: An interactive transformer for infrared and visible image fusion,'' \emph{Pattern Recognition}, vol. 156, p. 110822, 2024.

\bibitem{tang2022ydtr}
------, ``Ydtr: Infrared and visible image fusion via y-shape dynamic transformer,'' \emph{IEEE Transactions on Multimedia}, vol.~25, pp. 5413--5428, 2022.

\bibitem{tang2023datfuse}
W.~Tang, F.~He, Y.~Liu, Y.~Duan, and T.~Si, ``Datfuse: Infrared and visible image fusion via dual attention transformer,'' \emph{IEEE Transactions on Circuits and Systems for Video Technology}, vol.~33, no.~7, pp. 3159--3172, 2023.

\bibitem{hyeon2023scratching}
N.~Hyeon-Woo, K.~Yu-Ji, B.~Heo, D.~Han, S.~J. Oh, and T.-H. Oh, ``Scratching visual transformer's back with uniform attention,'' in \emph{Proceedings of the IEEE/CVF International Conference on Computer Vision}, 2023, pp. 5807--5818.

\bibitem{burt1987laplacian}
P.~J. Burt and E.~H. Adelson, ``The laplacian pyramid as a compact image code,'' in \emph{Readings in computer vision}.\hskip 1em plus 0.5em minus 0.4em\relax Elsevier, 1987, pp. 671--679.

\bibitem{zhan2017infrared}
L.~Zhan, Y.~Zhuang, and L.~Huang, ``Infrared and visible images fusion method based on discrete wavelet transform,'' \emph{Journal of Computers}, vol.~28, no.~2, pp. 57--71, 2017.

\bibitem{li2016infrared}
H.~Li, H.~Qiu, Z.~Yu, and Y.~Zhang, ``Infrared and visible image fusion scheme based on nsct and low-level visual features,'' \emph{Infrared Physics \& Technology}, vol.~76, pp. 174--184, 2016.

\bibitem{lu2014infrared}
X.~Lu, B.~Zhang, Y.~Zhao, H.~Liu, and H.~Pei, ``The infrared and visible image fusion algorithm based on target separation and sparse representation,'' \emph{Infrared Physics \& Technology}, vol.~67, pp. 397--407, 2014.

\bibitem{yang2020infrared}
Y.~Yang, Y.~Zhang, S.~Huang, Y.~Zuo, and J.~Sun, ``Infrared and visible image fusion using visual saliency sparse representation and detail injection model,'' \emph{IEEE Transactions on Instrumentation and Measurement}, vol.~70, pp. 1--15, 2020.

\bibitem{zhang2023joint}
C.~Zhang, H.~Li, Z.~Feng, and S.~He, ``Joint coupled dictionaries-based visible-infrared image fusion method via texture preservation structure in sparse domain,'' \emph{Computer Vision and Image Understanding}, vol. 235, p. 103781, 2023.

\bibitem{li2023infrared}
S.~Li, Y.~Zou, G.~Wang, and C.~Lin, ``Infrared and visible image fusion method based on principal component analysis network and multi-scale morphological gradient,'' \emph{Infrared Physics \& Technology}, vol. 133, p. 104810, 2023.

\bibitem{lu2014novel}
Y.~Lu, F.~Wang, X.~Luo, and F.~Liu, ``Novel infrared and visible image fusion method based on independent component analysis,'' \emph{Frontiers of Computer Science}, vol.~8, pp. 243--254, 2014.

\bibitem{zhang2014multi}
Y.~Zhang, L.~Chen, J.~Jia, and Z.~Zhao, ``Multi-focus image fusion based on non-negative matrix factorization and difference images,'' \emph{Signal Processing}, vol. 105, pp. 84--97, 2014.

\bibitem{cui2015detail}
G.~Cui, H.~Feng, Z.~Xu, Q.~Li, and Y.~Chen, ``Detail preserved fusion of visible and infrared images using regional saliency extraction and multi-scale image decomposition,'' \emph{Optics Communications}, vol. 341, pp. 199--209, 2015.

\bibitem{ma2017infrared}
J.~Ma, Z.~Zhou, B.~Wang, and H.~Zong, ``Infrared and visible image fusion based on visual saliency map and weighted least square optimization,'' \emph{Infrared Physics \& Technology}, vol.~82, pp. 8--17, 2017.

\bibitem{liu2022infrared}
Y.~Liu, L.~Dong, and W.~Xu, ``Infrared and visible image fusion via salient object extraction and low-light region enhancement,'' \emph{Infrared Physics \& Technology}, vol. 124, p. 104223, 2022.

\bibitem{goodfellow2014generative}
I.~Goodfellow, J.~Pouget-Abadie, M.~Mirza, B.~Xu, D.~Warde-Farley, S.~Ozair, A.~Courville, and Y.~Bengio, ``Generative adversarial nets,'' \emph{Advances in neural information processing systems}, vol.~27, 2014.

\bibitem{ma2020ddcgan}
J.~Ma, H.~Xu, J.~Jiang, X.~Mei, and X.-P. Zhang, ``Ddcgan: A dual-discriminator conditional generative adversarial network for multi-resolution image fusion,'' \emph{IEEE Transactions on Image Processing}, vol.~29, pp. 4980--4995, 2020.

\bibitem{vaswani2017attention}
A.~Vaswani, N.~Shazeer, N.~Parmar, J.~Uszkoreit, L.~Jones, A.~N. Gomez, {\L}.~Kaiser, and I.~Polosukhin, ``Attention is all you need,'' \emph{Advances in neural information processing systems}, vol.~30, 2017.

\bibitem{zhang2021transformer}
L.~Zhang and Y.~Wen, ``A transformer-based framework for automatic covid19 diagnosis in chest cts,'' in \emph{Proceedings of the IEEE/CVF international conference on computer vision}, 2021, pp. 513--518.

\bibitem{zhu2024towards}
Y.~Zhu, Y.~Ma, F.~Fan, J.~Huang, K.~Wu, and G.~Wang, ``Toward accurate infrared small target detection via edge-aware gated transformer,'' \emph{IEEE Journal of Selected Topics in Applied Earth Observations and Remote Sensing}, vol.~17, pp. 8779--8793, 2024.

\bibitem{lin2022ds}
A.~Lin, B.~Chen, J.~Xu, Z.~Zhang, G.~Lu, and D.~Zhang, ``Ds-transunet: Dual swin transformer u-net for medical image segmentation,'' \emph{IEEE Transactions on Instrumentation and Measurement}, vol.~71, pp. 1--15, 2022.

\bibitem{ma2022swinfusion}
J.~Ma, L.~Tang, F.~Fan, J.~Huang, X.~Mei, and Y.~Ma, ``Swinfusion: Cross-domain long-range learning for general image fusion via swin transformer,'' \emph{IEEE/CAA Journal of Automatica Sinica}, vol.~9, no.~7, pp. 1200--1217, 2022.

\bibitem{dai2024background}
\BIBentryALTinterwordspacing
Y.~Dai, M.~Xiao, Y.~Zhu, H.~Wang, K.~Guo, and J.~Yang, ``Background semantics matter: Cross-task feature exchange network for clustered infrared small target detection with sky-annotated dataset,'' 2024. [Online]. Available: \url{https://arxiv.org/abs/2407.20078}
\BIBentrySTDinterwordspacing

\bibitem{hu2018squeeze}
J.~Hu, L.~Shen, and G.~Sun, ``Squeeze-and-excitation networks,'' in \emph{Proceedings of the IEEE conference on computer vision and pattern recognition}, 2018, pp. 7132--7141.

\bibitem{zhu2021dau}
Y.~Zhu, S.~Tang, Y.~Jiang, and R.~Kang, ``Dau-net: A regression cell counting method,'' in \emph{ISCTT 2021; 6th International Conference on Information Science, Computer Technology and Transportation}.\hskip 1em plus 0.5em minus 0.4em\relax VDE, 2021, pp. 1--6.

\bibitem{huang2023rdca}
Z.~Huang, B.~Yang, and C.~Liu, ``Rdca-net: Residual dense channel attention symmetric network for infrared and visible image fusion,'' \emph{Infrared Physics \& Technology}, vol. 130, p. 104589, 2023.

\bibitem{liu2022ssau}
S.~Liu, S.~Liu, S.~Zhang, B.~Li, W.~Hu, and Y.-D. Zhang, ``Ssau-net: A spectral--spatial attention-based u-net for hyperspectral image fusion,'' \emph{IEEE Transactions on Geoscience and Remote Sensing}, vol.~60, pp. 1--16, 2022.

\bibitem{wang2023attention}
Z.~Wang, Z.~Wu, X.~Li, H.~Shao, T.~Han, and M.~Xie, ``Attention-aware temporal--spatial graph neural network with multi-sensor information fusion for fault diagnosis,'' \emph{Knowledge-Based Systems}, vol. 278, p. 110891, 2023.

\bibitem{liu2021swin}
Z.~Liu, Y.~Lin, Y.~Cao, H.~Hu, Y.~Wei, Z.~Zhang, S.~Lin, and B.~Guo, ``Swin transformer: Hierarchical vision transformer using shifted windows,'' in \emph{Proceedings of the IEEE/CVF international conference on computer vision}, 2021, pp. 10\,012--10\,022.

\bibitem{woo2018cbam}
S.~Woo, J.~Park, J.-Y. Lee, and I.~S. Kweon, ``Cbam: Convolutional block attention module,'' in \emph{Proceedings of the European conference on computer vision (ECCV)}, 2018, pp. 3--19.

\bibitem{qin2021fcanet}
Z.~Qin, P.~Zhang, F.~Wu, and X.~Li, ``Fcanet: Frequency channel attention networks,'' in \emph{Proceedings of the IEEE/CVF international conference on computer vision}, 2021, pp. 783--792.

\bibitem{zagoruyko2016paying}
S.~Zagoruyko and N.~Komodakis, ``Paying more attention to attention: Improving the performance of convolutional neural networks via attention transfer,'' \emph{arXiv preprint arXiv:1612.03928}, 2016.

\bibitem{shreyamsha2015image}
B.~Shreyamsha~Kumar, ``Image fusion based on pixel significance using cross bilateral filter,'' \emph{Signal, image and video processing}, vol.~9, pp. 1193--1204, 2015.

\bibitem{li2018infrared}
H.~Li and X.-J. Wu, ``Infrared and visible image fusion using latent low-rank representation,'' \emph{arXiv preprint arXiv:1804.08992}, 2018.

\bibitem{zhang2020rethinking}
H.~Zhang, H.~Xu, Y.~Xiao, X.~Guo, and J.~Ma, ``Rethinking the image fusion: A fast unified image fusion network based on proportional maintenance of gradient and intensity,'' in \emph{Proceedings of the AAAI conference on artificial intelligence}, vol.~34, no.~07, 2020, pp. 12\,797--12\,804.

\bibitem{xu2021classification}
H.~Xu, H.~Zhang, and J.~Ma, ``Classification saliency-based rule for visible and infrared image fusion,'' \emph{IEEE Transactions on Computational Imaging}, vol.~7, pp. 824--836, 2021.

\bibitem{yang2024mda}
G.~Yang, J.~Li, H.~Lei, and X.~Gao, ``A multi-scale information integration framework for infrared and visible image fusion,'' \emph{Neurocomputing}, vol. 600, p. 128116, 2024.

\bibitem{jia2021llvip}
X.~Jia, C.~Zhu, M.~Li, W.~Tang, and W.~Zhou, ``Llvip: A visible-infrared paired dataset for low-light vision,'' in \emph{Proceedings of the IEEE/CVF international conference on computer vision}, 2021, pp. 3496--3504.

\bibitem{toet2017tno}
A.~Toet, ``The tno multiband image data collection,'' \emph{Data in brief}, vol.~15, pp. 249--251, 2017.

\bibitem{tang2022piafusion}
L.~Tang, J.~Yuan, H.~Zhang, X.~Jiang, and J.~Ma, ``Piafusion: A progressive infrared and visible image fusion network based on illumination aware,'' \emph{Information Fusion}, vol.~83, pp. 79--92, 2022.

\bibitem{xu2020fusiondn}
H.~Xu, J.~Ma, Z.~Le, J.~Jiang, and X.~Guo, ``Fusiondn: A unified densely connected network for image fusion,'' in \emph{Proceedings of the AAAI conference on artificial intelligence}, vol.~34, no.~07, 2020, pp. 12\,484--12\,491.

\bibitem{brown2011multi}
M.~Brown and S.~S{\"u}sstrunk, ``Multi-spectral sift for scene category recognition,'' in \emph{CVPR 2011}.\hskip 1em plus 0.5em minus 0.4em\relax IEEE, 2011, pp. 177--184.

\bibitem{qu2002information}
G.~Qu, D.~Zhang, and P.~Yan, ``Information measure for performance of image fusion,'' \emph{Electronics letters}, vol.~38, no.~7, p.~1, 2002.

\bibitem{wang2005nonlinear}
Q.~Wang, Y.~Shen, and J.~Q. Zhang, ``A nonlinear correlation measure for multivariable data set,'' \emph{Physica D: Nonlinear Phenomena}, vol. 200, no. 3-4, pp. 287--295, 2005.

\bibitem{xydeas2000objective}
C.~S. Xydeas, V.~Petrovic \emph{et~al.}, ``Objective image fusion performance measure,'' \emph{Electronics letters}, vol.~36, no.~4, pp. 308--309, 2000.

\bibitem{zhao2007performance}
J.~Zhao, R.~Laganiere, and Z.~Liu, ``Performance assessment of combinative pixel-level image fusion based on an absolute feature measurement,'' \emph{Int. J. Innov. Comput. Inf. Control}, vol.~3, no.~6, pp. 1433--1447, 2007.

\bibitem{li2008novel}
S.~Li, R.~Hong, and X.~Wu, ``A novel similarity based quality metric for image fusion,'' in \emph{2008 International Conference on Audio, Language and Image Processing}.\hskip 1em plus 0.5em minus 0.4em\relax IEEE, 2008, pp. 167--172.

\bibitem{han2013new}
Y.~Han, Y.~Cai, Y.~Cao, and X.~Xu, ``A new image fusion performance metric based on visual information fidelity,'' \emph{Information fusion}, vol.~14, no.~2, pp. 127--135, 2013.

\bibitem{yolov5}
\BIBentryALTinterwordspacing
G.~Jocher, ``Ultralytics yolov5,'' 2020. [Online]. Available: \url{https://github.com/ultralytics/yolov5}
\BIBentrySTDinterwordspacing

\end{thebibliography}

\end{document}